\documentclass[acmlarge]{acmart}
\usepackage{subfigure}
\usepackage{color}
\usepackage{amsmath}

\newcommand{\para}[1]{{\vspace{4pt} \bf \noindent #1 \hspace{5pt}}}

\theoremstyle{definition}
\newtheorem{defn}{\hspace{2em}Definition} %

\usepackage{algorithmic}
\usepackage[ruled]{algorithm2e} 

\definecolor{codegreen}{rgb}{0,0.6,0}
\definecolor{codegray}{rgb}{0.5,0.5,0.5}
\definecolor{codepurple}{rgb}{0.58,0,0.82}
\definecolor{backcolour}{rgb}{1,1,1}
\definecolor{codeblack}{rgb}{0,0,0}
\definecolor{b}{rgb}{0,0,0}
\definecolor{nb}{rgb}{0,0,0}
\definecolor{nb2}{rgb}{0,0,0}

\AtBeginDocument{%
  \providecommand\BibTeX{{%
    \normalfont B\kern-0.5em{\scshape i\kern-0.25em b}\kern-0.8em\TeX}}}

\begin{document}

\title{	"How do urban incidents affect traffic speed?" A Deep Graph Convolutional Network for Incident-driven Traffic Speed Prediction}


\author{Qinge Xie}
\affiliation{%
  \institution{School of Computer Science, Fudan University}
}
\email{qgxie17@fudan.edu.cn}

\author{Tiancheng Guo}
\affiliation{%
  \institution{School of Computer Science, Fudan University}
}
\email{tcguo16@fudan.edu.cn}

\author{Yang Chen}
\affiliation{%
  \institution{School of Computer Science, Fudan University}
}
\email{chenyang@fudan.edu.cn}

\author{Yu Xiao}
\affiliation{%
  \institution{Department of Communications and Networking, Aalto University}
}
\email{yu.xiao@aalto.fi}

\author{Xin Wang}
\affiliation{%
	\institution{School of Computer Science, Fudan University}
}
\email{xinw@fudan.edu.cn}

\author{Ben Y. Zhao}
\affiliation{%
	\institution{Computer Science, University of Chicago}
}
\email{ravenben@cs.uchicago.edu}


\begin{abstract}
Accurate traffic speed prediction is an important and challenging topic for transportation planning. Previous studies on traffic speed prediction predominately used spatio-temporal and context features for prediction. However, they have not made good use of the impact of urban traffic incidents. In this work,  we aim to make use of the information of urban incidents to achieve a better prediction of traffic speed. Our incident-driven prediction framework consists of three processes. First, we propose a critical incident discovery method to discover urban traffic incidents with high impact on traffic speed. Second, we design a binary classifier, which uses deep learning methods to extract the latent incident impact features from the middle layer of the classifier. Combining above methods, we propose a Deep Incident-Aware Graph Convolutional Network (DIGC-Net) to effectively incorporate urban traffic incident, spatio-temporal, periodic and context features for traffic speed prediction. We conduct experiments on two real-world urban traffic datasets of San Francisco and New York City. The results demonstrate the superior performance of our model compare to the competing benchmarks.
\end{abstract}



\maketitle

\section{Introduction}
\label{introduction}

Traffic speed prediction has been a challenging problem for decades, and has a wide range of traffic planning and related applications, including congestion control~\cite{li2017reinforcement}, vehicle routing planning~\cite{johnson2017beautiful}, urban road planning~\cite{rathore2016urban} and travel time estimation~\cite{gao2019aggressive}. The difficulty of the problem comes from the complex and highly dynamic nature of traffic and road conditions, as well as a variety of other unpredictable, ad hoc factors. Urban traffic incidents, including lane restriction, road construction and traffic collision, which is one of the most important factors, tend to dramatically impact traffic for limited time periods. Yet the frequency of these events means their aggregate impact cannot be ignored when modeling and predicting traffic speed. 

Despite a large amount of research on detecting urban traffic incidents~\cite{Detect_Anomalies_ubicomp18,convlstm_KDD18}, a small number of works study the impact of urban traffic incidents recently. \cite{INCI_URB12} proposed a system for predicting the cost and impact of highway incidents. \cite{javid2018framework} developed a framework to estimate travel time variability caused by traffic incidents. \cite{he2019traffic} proposed to use the ratio of speed before and after incident as the traffic impact coefficient to evaluate the traffic influence of an incident. Those works have proven the significant impact of urban traffic incidents on traffic conditions.
However, improving traffic speed prediction by traffic incidents has not been well explored. {\color{nb2}Some previous works~\cite{lin2017road} use incident data collected from social networks (e.g., Twitter) by keywords to improve traffic prediction. However, they fail to consider the impact level of different urban traffic incidents but treat all incidents equally for speed prediction.} Today, the large majority solutions including traditional machine learning~\cite{castro2009online}, matrix decomposition~\cite{LSMRN_kdd16} and deep learning methods~\cite{DCRNN_ICLR18,LCRNN_ijcai18,yao2019revisiting} of traffic speed prediction mainly use spatio-temporal features of traffic network and context features such as weather data. These solutions for predicting traffic speed do not factor in the impact of those dynamic traffic incidents. 

A number of questions naturally arise: how do different {\color{nb2}urban} traffic incidents impact traffic flow speeds? Do high impact traffic incidents have specific spatio-temporal patterns {\color{nb2}in the city}? How can we use {\color{nb2}urban} traffic incident data to improve traffic speed prediction? In this paper, our goal is to answer these questions, and in doing so, understand the impact of {\color{nb2}urban} traffic incidents on traffic speeds and propose an effective framework using {\color{nb2}urban} traffic incident information to improve traffic speed prediction. There are two main challenges in our incident-driven traffic speed prediction problem. First, the impact of {\color{nb2}urban} traffic incidents is complex and varies significantly across incidents. For example, incidents occur in the wee hours and in remote areas will have little impact on adjacent roads, while incidents during the rush hours and in high-traffic areas (e.g. downtown) are very likely to affect the surrounding traffic flows or even cause congestion~\cite{RUSH_ICDM12}. Therefore, it is unreasonable to treat all {\color{nb2}urban} traffic incidents equally for traffic speed prediction, which may even negatively impact prediction performance. Second, the impact of {\color{nb2}urban} traffic incidents on adjacent roads will be affected by external factors like incident occurrence time, incident type and road topology structure. We need to extract the latent impact features of traffic incidents on traffic flows to improve traffic speed prediction.

To tackle the first challenge, we propose a critical incident discovery method to quantify the impact of {\color{nb2}urban} traffic incidents on traffic flows. We consider both anomalous degree and speed variation of adjacent roads to discover the critical traffic incidents. Next, to tackle the second challenge, we propose a binary classifier which uses deep learning methods to extract the latent impact features of incidents. The impact of incidents varies in degree and the impact is neither binary nor strict multi-class. So we extract the latent impact features from the middle layer of the classifier, where the latent features are continuous and filtered. We adopt Graph Convolution Network (GCN)~\cite{GCN_ICLR14} to capture spatial features of road networks. GCN is known to be able to effectively capture the topology features in non-Euclidean structures and the complex road network is a typical non-Euclidean structure. Combining above methods, we propose a Deep Incident-Aware Graph Convolutional Network (DIGC-Net) to improve traffic prediction by traffic incident data. DIGC-Net can effectively leverage traffic incident, spatio-temporal, periodic and context features for prediction. 

We test our framework using two real-world {\color{nb2}urban} traffic datasets from San Francisco and New York City. Experimental results empirically answer the above mentioned questions, and also show the particularly different spatio-temporal distributions of critical/non-critical incidents. We compare DIGC-Net with state-of-the-art methods, and the results demonstrate the superior performance of our model and also verify that the incident learning component is the key to the improvement of prediction performance.

\begin{figure}[t]
	\centering
	\centering
	\subfigure[Road network of SFO]{
		\label{fig:rn:a} 
		\includegraphics[width=1.4in,height=1.2in]{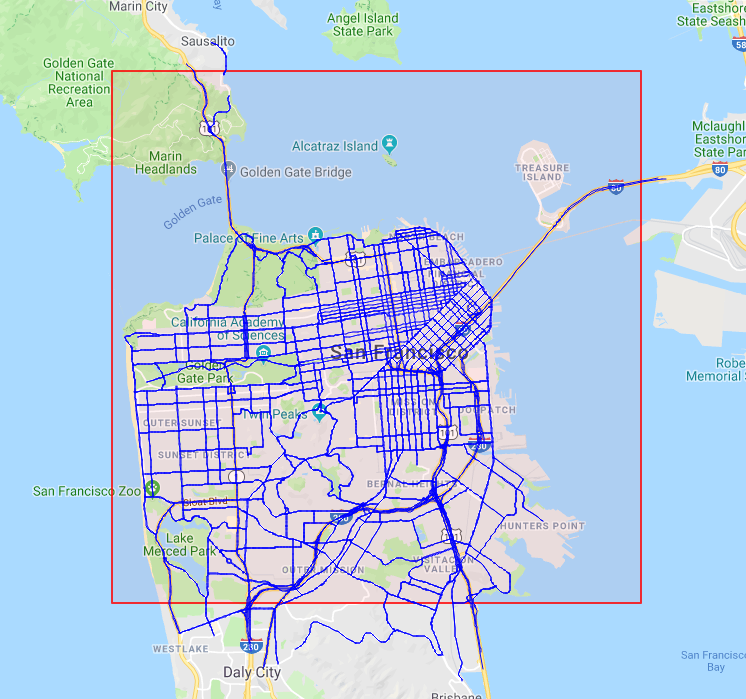}}
	\hspace{0.2in}
	\subfigure[The congestion incident]{
		\label{fig:example1:b} 
		\includegraphics[width=1.4in,height=1.2in]{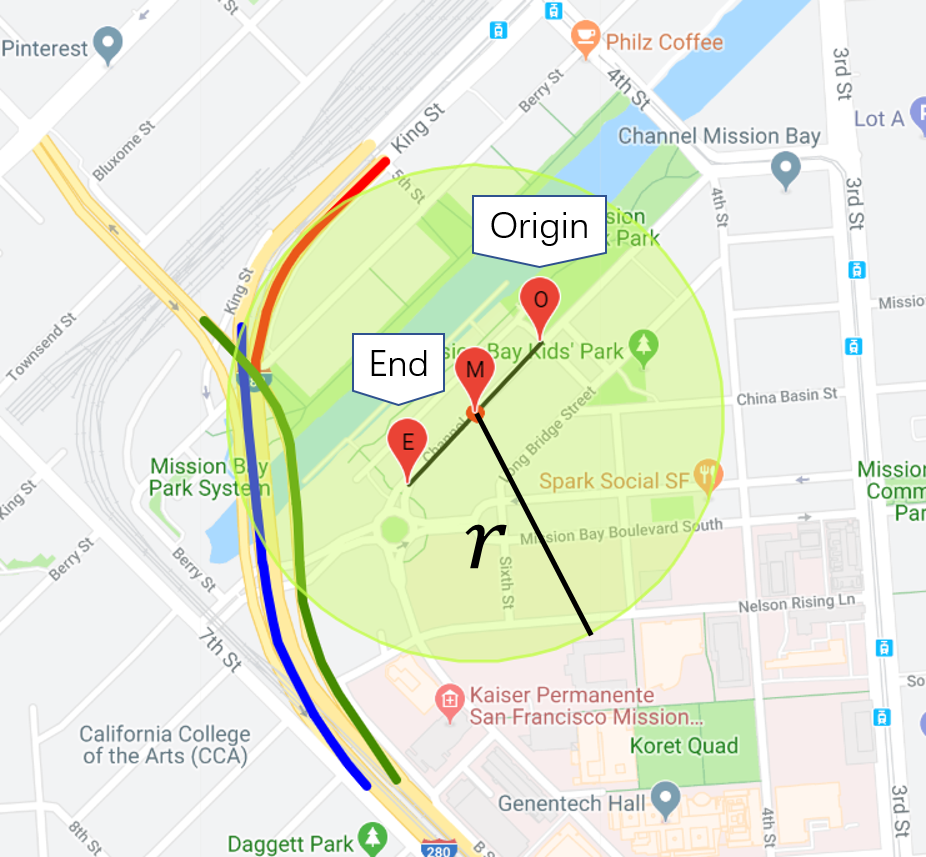}}
	\hspace{0.2in}
	\subfigure[Speed curves of three candidate flows]{
					\label{fig:example1:c} 
					\includegraphics[width=2in,height=1.2in]{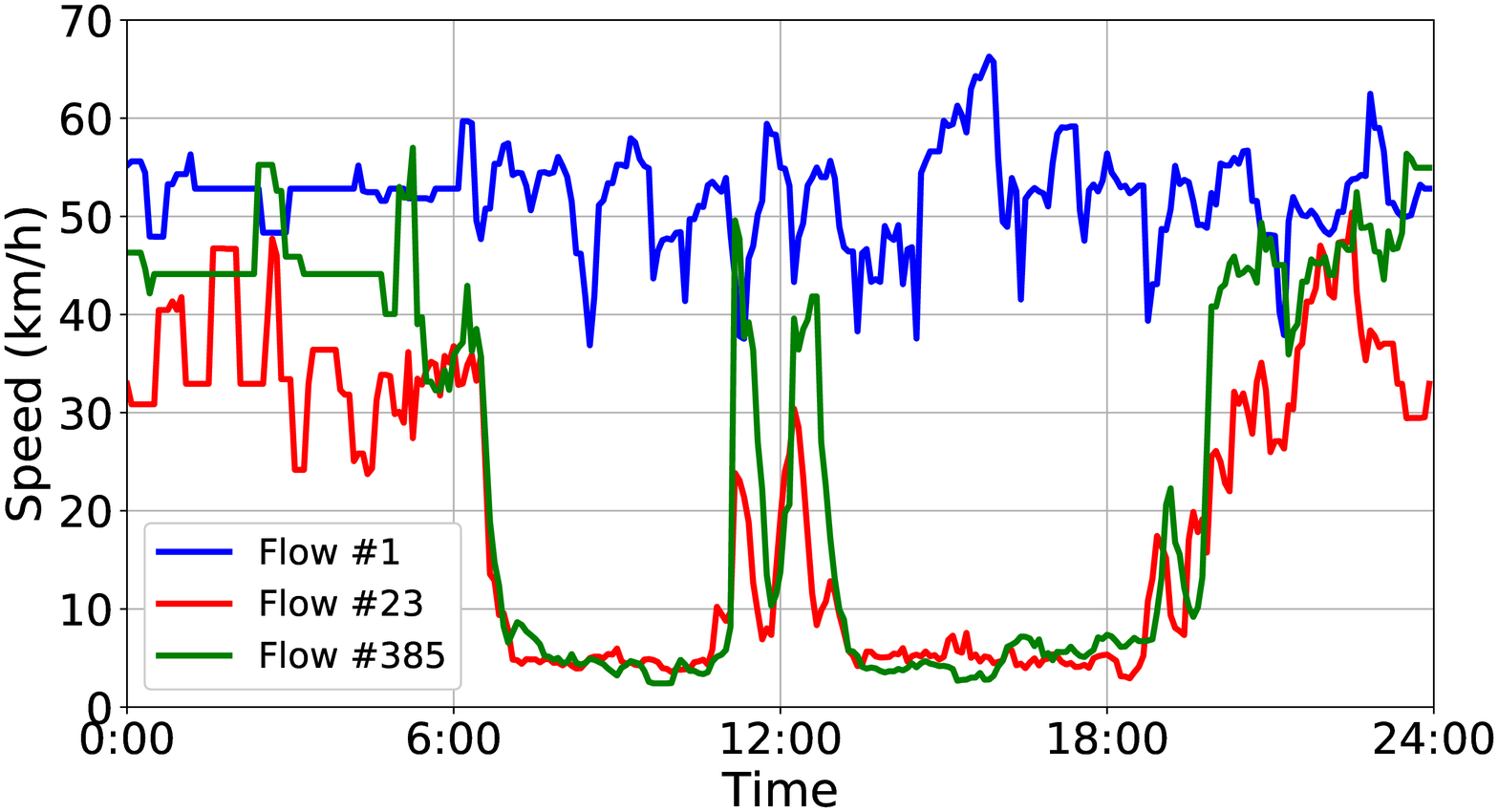}  }
	\caption{{\color{nb2}Traffic illustration of SFO}}
	\label{fig:rn} 
	\vspace{-0.1in}
\end{figure}

We summarize our key contributions as follows:
\begin{itemize}
	\item To quantify the impact of traffic incidents on traffic speeds, we propose a critical incident discover method and discover critical incidents in the city. We further explore the spatio-temporal distributions of critical/non-critical incidents and find noteworthy differences.
	\item In order to extract the latent incident impact features, we skillfully design a binary classifier to extract the latent impact features from the middle layer of the classifier. We use the binary classifier as an internal component of our final framework to improve traffic speed prediction. 
	\item  We propose a DIGC-Net to effectively incorporate incident, spatio-temporal, periodic and context features for traffic speed prediction. We conduct experiments using two real-world urban traffic datasets, and results show that DIGC-Net outperforms competing benchmarks and {\color{nb2} the incident learning component is the key to the improvement of prediction performance. Meanwhile, the incident learning component can be flexibly insert to other models as a common use to learning incident impact features}.
	
\end{itemize}

\section{Preliminaries}
\label{overview}
Before diving into details of the model, we begin with some preliminaries on our datasets and problem formulation in this section. 

\subsection{Datasets}

We utilize two datasets, a traffic dataset and an attribute dataset (weather data). The traffic dataset consists of traffic road network, speed and incident sub-dataset from two major metropolitan areas, San Francisco (SFO) and New York City (NYC), with complex traffic conditions and varying physical features that may affect latent traffic patterns~\cite{City_Behavior_IEEE18}. We collect the weather dataset using Yahoo Weather API~\cite{Yahoo_Weather} and fields includes weather type, temperature and sunrise time. We collect the traffic dataset from a public API: HERE Traffic~\cite{Here}. 1) Road Network: We set lat/lng bounding boxes (Figure~\ref{fig:rn:a}) on two cities of SFO (37.707,-122.518/37.851,-122.337) and NYC (40.927,-74.258/40.495,-73.750) 
to gather the internal road network. 2) Traffic Speed: We collect the real-time traffic speed of each flow in the areas described above and record real-time speeds of each flow every 5 minutes. 3) Traffic Incident: We also collect the traffic incident data in same areas every 5 minutes. For each incident, we can get the incident features like type and location. 

\begin{table}[t]%
	\centering
	\caption{Overall datasets}
	\begin{tabular}{|p{4cm} |p{8cm}|}
		\toprule
		Component             &       Datasets Description       \\
		\midrule[1pt]
		Critical Incident Discovery      & Use traffic incident, road network and speed sub-dataset. The incident and speed data are from Apr. 17 to Apr. 24, 2019.  \\
		\midrule[1pt]
		Impact Features Extraction      &  Use traffic incident, road network and speed sub-dataset. The incident and speed data are from Apr. 17 to Apr. 24, 2019.  \\
		\midrule[1pt]
		DIGC-Net      & Use traffic incident, road network, speed sub-dataset and weather
		dataset.  The incident and speed data are
		form Apr. 4 to May 2, 2019 (4 weeks).\\
		\bottomrule
	\end{tabular}
	\label{tab:datasets}
\end{table}%


{\color{nb}\para{Flow.} The real-time speeds in different segments of one single road are discrete. HERE divides every road into multiple segments. We denote one road segment as one flow $\xi$. Every flow at each time slot will have a speed and we use flow as the smallest unit in the road network. }

\subsection{Problem Formulation and Preprocessing}
\label{section_pf}
First, we denote a road network as an undirected graph $\mathcal N=\left(V,E\right)$, where each node represents an intersection or a split point on the road, and each edge represents a road segment.

\para{Reconstruction of the road network.} As our task is to predict the speed of every road segment, we use the road segment as the node. More specifically, we use every flow as one node to build the road network. If two flows $\xi_i$ and $\xi_j$ have points of intersection, we will add an edge to connect node $\xi_i$ and node $\xi_j$. Therefore, we build a new road network graph $\mathcal G=\left(V,E\right)$, where each node represents a flow and each edge represents an intersection of the flows or a split point on the flow. There are 2,416 nodes and 19,334 edges of SFO, and 13,028 nodes and 92,470 edges of NYC. We will use the re-build road network graph $\mathcal G$ in the rest of the paper.

\para{Problem formulation.} We use $v^t_{\xi_i}$ to represent the speed of flow $\xi_i$ at time slot $t$. For every speed snapshot of the road network, we will get a vector of all flows $V^t = \left[v^t_{\xi_0},v^t_{\xi_1},\cdots,v^t_{\xi_{N-1}}   \right]$, where $N$ is the total number of flows. Given the re-build road graph $\mathcal G=\left(V, E\right)$ and a T-length historical real-time speed sequence $[V^{t-T}, V^{t-T+1}, \cdots, V^{t-1}]$ of all flows, our task is to predict future speeds of every flow in the city, \textit{i.e.}, $Y=\left[V^{t}, V^{t+1}, \cdots, V^{t+k-1}\right]$, where $k$ is the prediction length. Given a set of urban traffic incidents occur close to the predicted time $t$, more specifically, a set of incidents occur within $[t-T_1, t-T_2]$, where $t-T_1$ is the earliest included incident occurrence time and $t-T_2$ is the latest included incident occurrence time. We extract the features of the impact of above mentioned incidents on traffic flows to improve the speed prediction performance.

\section{Urban Critical Incident Discovery}

\label{sec:incident}

The impact of urban traffic incidents are complex and also influenced by other factors like {\color{nb2}the topological structure of urban road network, temporal features} and incident type.
Treating all urban traffic incidents equally will add additional noise to traffic speed prediction process. 
In this section, we focus on analyzing the impact of different urban traffic incidents, and introduce our urban critical incident discovery methodology.

\subsection{Methodology}

\para{Case Study: A Congestion Incident. } Figure~\ref{fig:example1:b} presents a congestion incident occurred at 06:32 am on Apr. 17, 2019 in San Francisco. $M$ is the center point of the incident and we set $r$ to represent the radius of the impact range. The circle with the center $M$ and radius $r$ stands for the region affected by the incident. We define that if the center of flow is in the circle, then the flow might be affected by the incident. The circle in Figure~\ref{fig:example1:b} presents the affected region when $r=300\,m$. The blue, red and green lines represent three flows $\xi_1$, $\xi_{23}$ and $\xi_{385}$ in San Francisco which might be affected by the incident, respectively. The speed curves of the three candidate flows are shown in Figure~\ref{fig:example1:c}. We observe that during 6:00 am - 7:00 am, the speeds of $\xi_{23}$ and $\xi_{385}$ show a sharp reduction while the variation of $\xi_{1}$ is relatively slight, but it still become more choppy after the incident occurred.

Next, we analyze each candidate flow that whether it will truly be affected by the incident. We use a variant of the method proposed in~\cite{Detect_Anomalies_ubicomp18} to compute the anomalous degree of each flow. {\color{nb2}They divides the city area into several grids and compute the anomalous degree of each grid region to detect urban anomalies.} The key idea to compute the anomalous degree of a region is based on its historically similar regions in the city. The sudden drop of speed similarity of a region and its historically similar regions indicates the occurrence of urban anomalies, {\color{nb2}and the well-designed experiments had verified the effectiveness of the detection method}. In our problem, we use each flow as the unit rather than grid region.

\begin{defn}{\bfseries{Pair-wise Similarity of Flows.}}
Given two flows at time slot $t$ with speeds $v^t_{\xi_i}$ and $v^t_{\xi_j}${\color{nb}, for a time window }$W = \left[t-T+1:t\right]$, the pair-wise similarity is calculated by:
	\begin{equation} \label{eq_P}
	s^{[t-T+1:t]}_{{\xi_i},{\xi_j}} = P\left(v^{[t-T+1:t]}_{\xi_i},v^{[t-T+1:t]}_{\xi_j} \right),
	\end{equation}
	
	where $P$ is to calculate Pearson correlation coefficient~\cite{Pearson} of {\color{nb}two speed sequences}. Then the similarity matrix $S$ of all flows at $t$ is calculated by the following equation:
	 
		\begin{equation} \label{eq_S}
		S^t = \begin{vmatrix} s^{[t-T+1:t]}_{{\xi_0},{\xi_0}}  & \cdots\ &  s^{[t-T+1:t]}_{{\xi_0},{\xi_N-1}}\\ 
		
		\cdots\ & \ddots\ & \cdots\ \\
		s^{[t-T+1:t]}_{{\xi_N-1},{\xi_0}} &  \cdots\ & s^{[t-T+1:t]}_{{\xi_N-1},{\xi_N-1}}
		
		\end{vmatrix},
		\quad
	\end{equation}
	
	where $N$ is the total number of flows in the city.
\end{defn}

\begin{defn}{\bfseries{Similarity Decrease Matrix (SD).}} Similar to~\cite{Detect_Anomalies_ubicomp18}, we define the similarity decrease matrix $SD$, which represents the decreased similarity of each flow pair from time slot $t-1$ to $t$.  $SD$ at time slot $t$ is calculated by: $SD^t = max\left(0, S^{t-1} - S^{t}\right)$. Zeroing the numbers less than zero is due to that we only consider the case where the similarity goes down.
\end{defn}

\begin{defn}{\bfseries{Anomalous Degree (AD).}} Then we use similarity matrix $S$ and similarity decrease matrix $SD$ to compute $AD$ of flows at time slot $t$. We use a threshold parameter $\delta$ to capture the historically similar flows. When the similarity of two flows is equal or greater than $\delta$, we define they are historically similar. Given a flow $\xi_i$ at time slot $t$, the historically similar flow sets of $\xi_i$ is denoted as $HS^{t}_{\xi_i} =  \{ \xi_j \mid i \not= j \, and \, 
	S^{t}_{i,j}=S^{t}_{j,i}  \geq \delta \}$. {\color{nb2} Pair-wise similarity is computed by Pearson correlation coefficient (PCC) and PCC in [0.5, 0.7] indicates variables are moderately correlated according to~\cite{rumsey2015u}. Therefore, we set $\delta=0.5$ here to select the historically similar flows which are at least moderately similarity to the flow $\xi_i$. } Anomalous degree of flow $\xi_i$  at time slot $t$ is calculated by the following equation:
	
			\begin{equation} \label{eq_SD}
	AD^t_{\xi_i} = \frac{\Sigma_{\xi_j \in HS^t_{\xi_i}}S^{t-1}_{i,j} \cdot SD^{t}_{i,j} }{\Sigma_{\xi_j \in HS^t_{\xi_i}}S^{t-1}_{i,j} },
	\end{equation}
	
	{\color{nb} where $AD$ is the decrease degree in speed similarity of $\xi_i$ and its historically similar flows.}
\end{defn}

\begin{figure}[t]
		\centering
		\subfigure[Anomalous degree]{
			\label{fig:me:a} 
			\includegraphics[width=2.5in,height=1.5in]{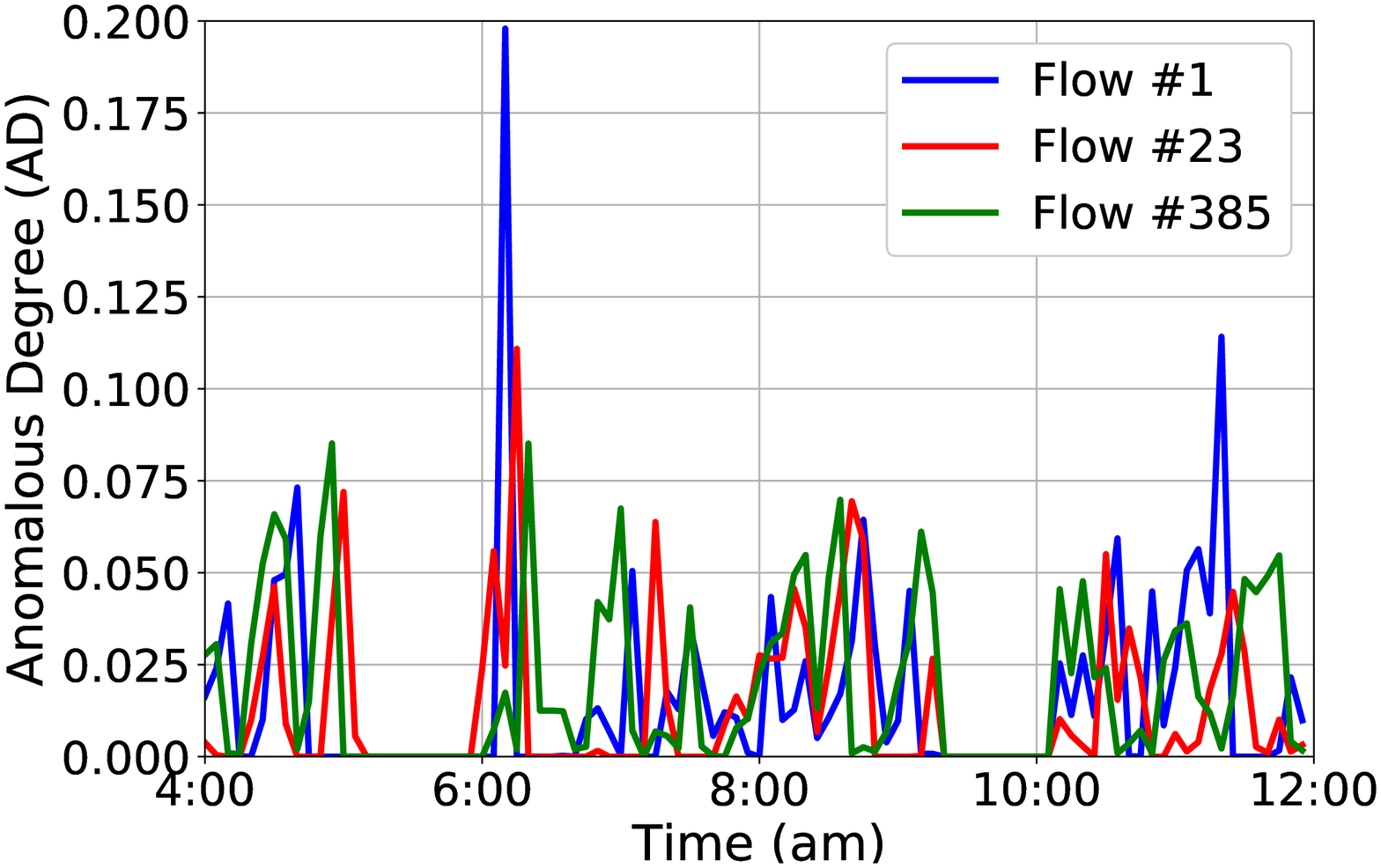}}
		\hspace{0.4in}
		\subfigure[{\color{b}Relative speed variation rate}]{
			\label{fig:me:b} 
			\includegraphics[width=2.5in,height=1.5in]{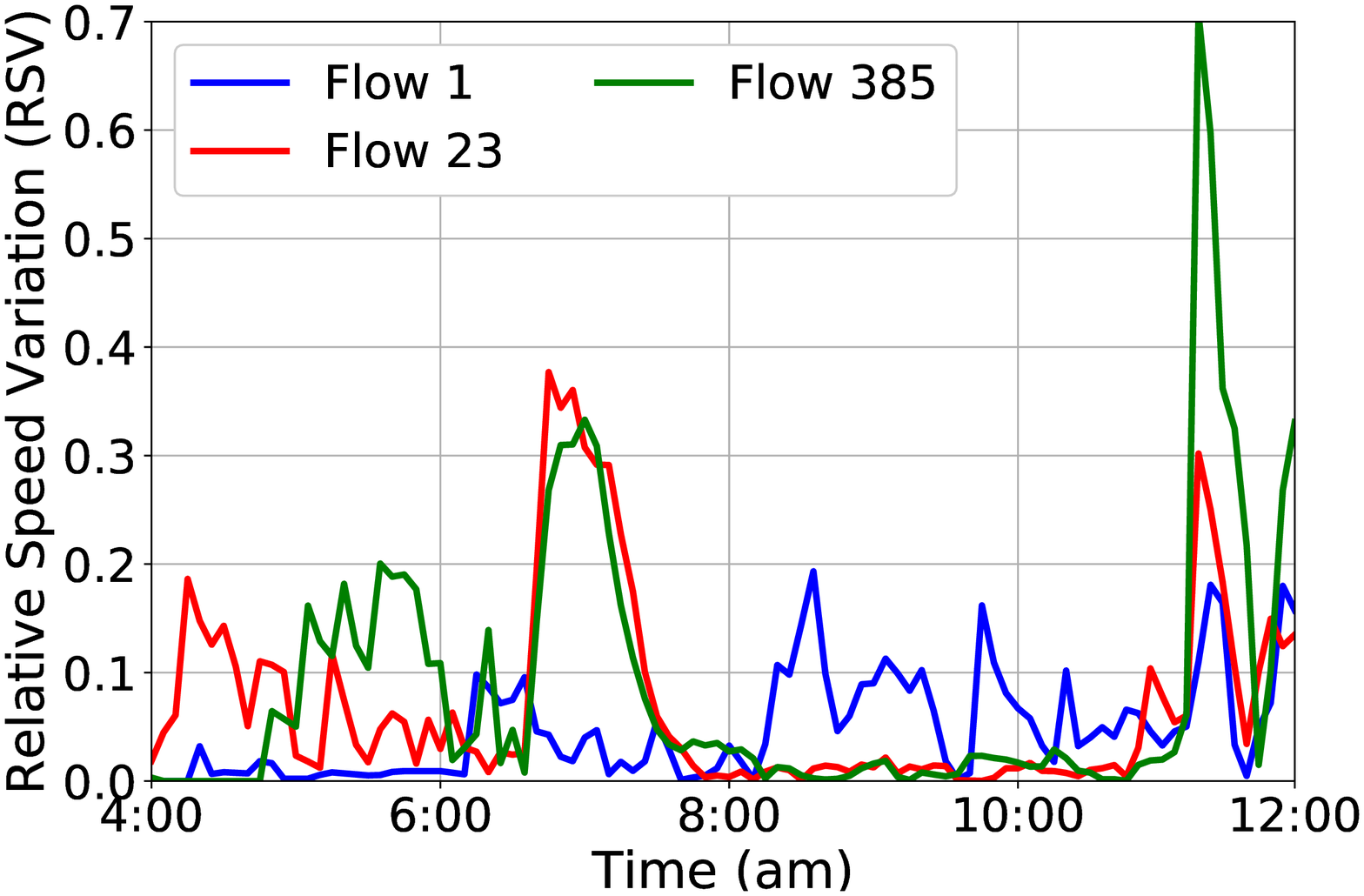}}
		\caption{{\color{nb}$AD$ and $RSV$ of three candidate flows}}
		\label{fig:me} 
	
\end{figure}

\begin{figure}[t]  
	\centering  
	\begin{minipage}{0.3\linewidth} 
		\setlength{\abovecaptionskip}{0.3in}  
		\centering      
		\includegraphics[width=1.5in, height=1.2in]{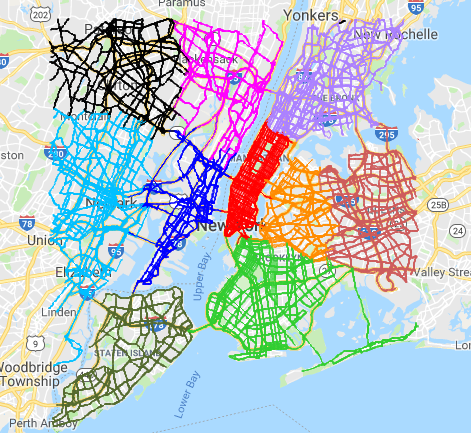}
		\caption{Clusters of NYC}
		\label{fig:clu} 
	\end{minipage}   
	\begin{minipage}{0.6\linewidth} 
		\centering
		\setlength{\abovecaptionskip}{-0.05in} 
		\subfigure[SFO]{
			\label{fig:inci_res:a} 
			\includegraphics[width=1.7in,height=1.2in]{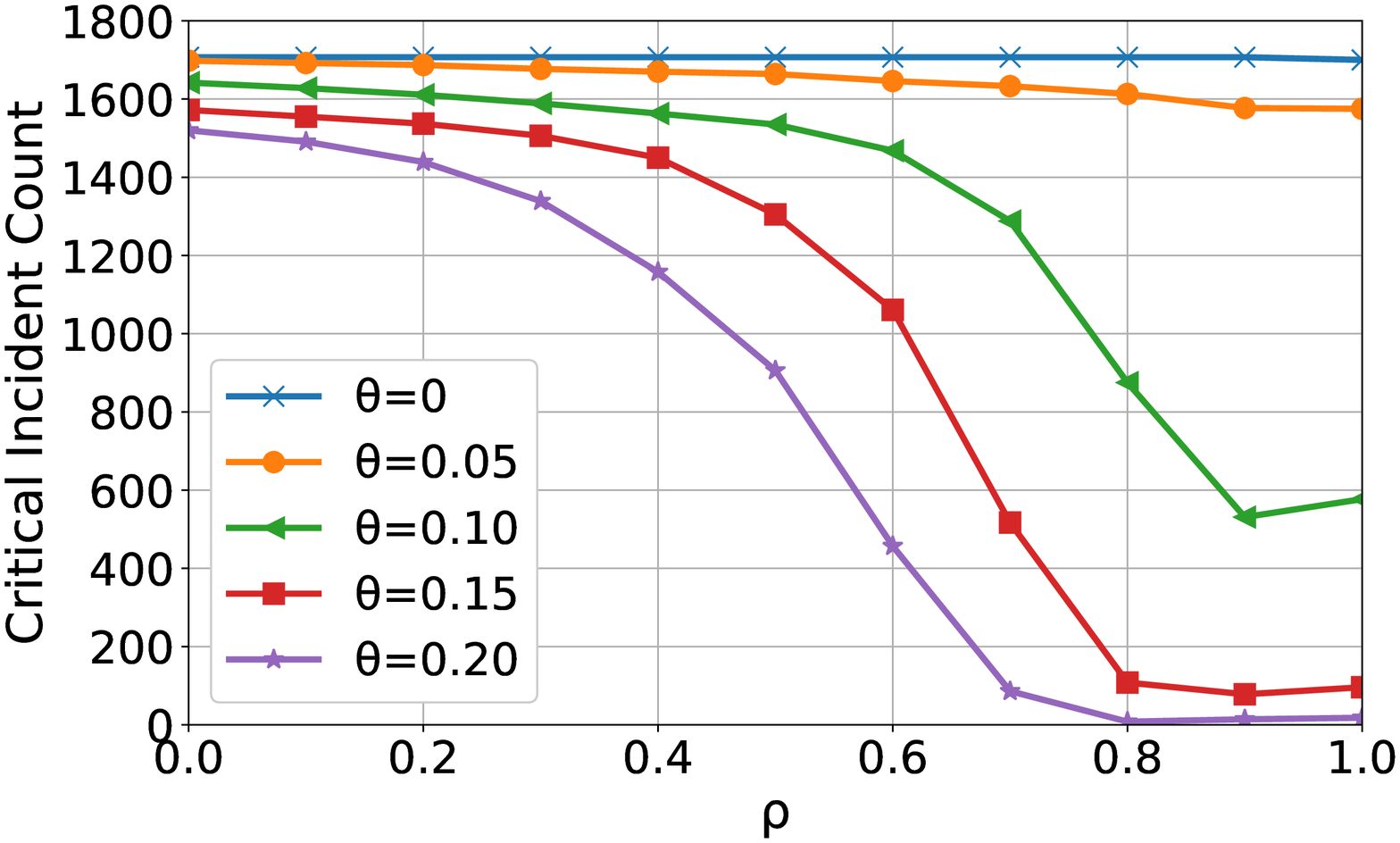}}
		\hspace{0.15in}
		\subfigure[{NYC}]{
			\label{fig:inci_res:b} 
			\includegraphics[width=1.7in,height=1.2in]{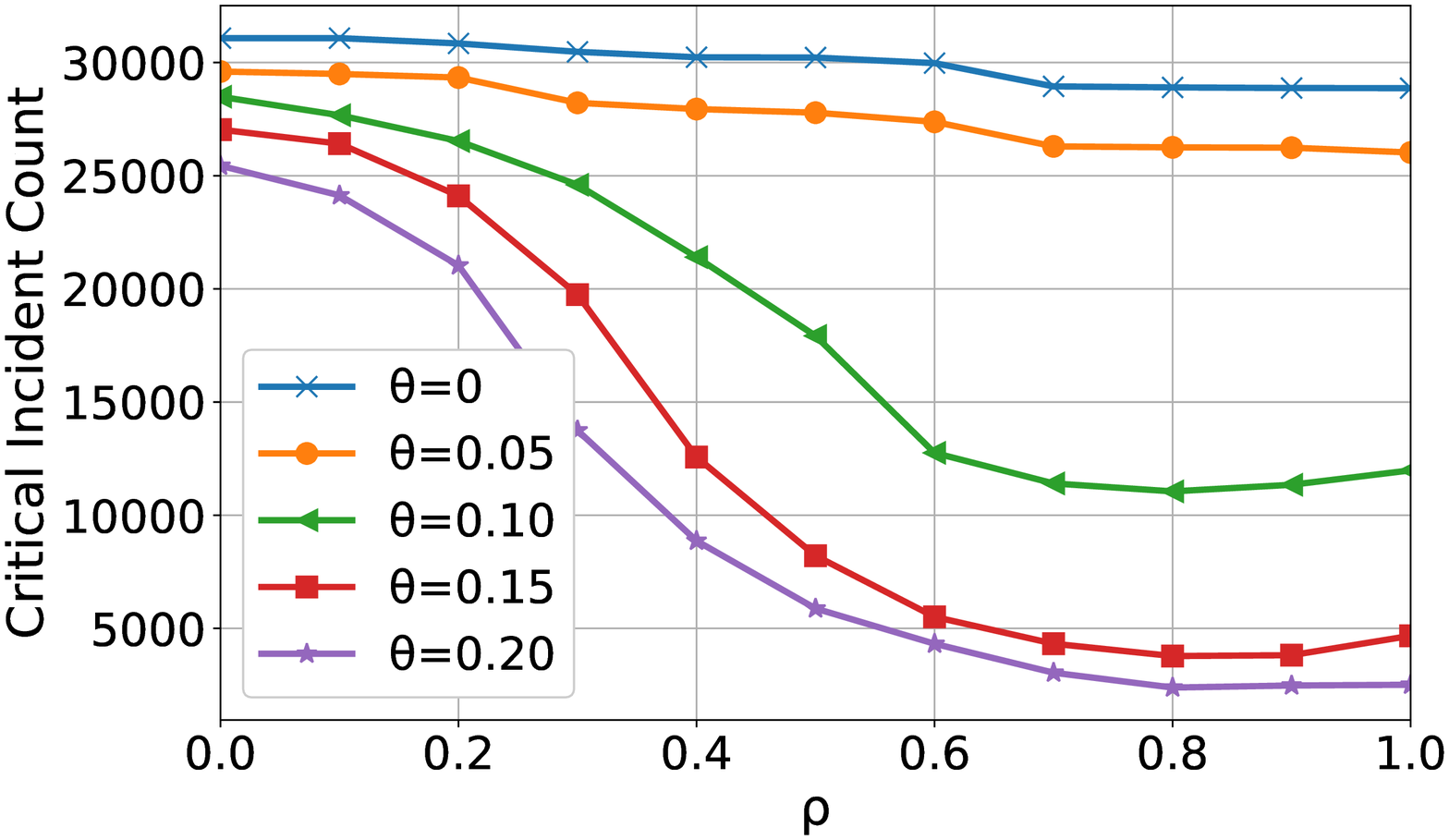}}
		\caption{Varying $\rho$ and $\theta$}
		\label{fig:inci_res} 
	\end{minipage}%
\end{figure}

{\color{nb2}
\para{Local Anomalous Degree Algorithm. } The time complexity of computing similarity matrix $S$ is $\mathcal{O} \left( N^2 \dot T \right)$, where $N$ is the number of flows and $T$ is the length of historical speed sequences. For cities with complex traffic road networks such as New York City (13,028 flows), it will cost a lot to compute similarity matrix $S$, similarity decrease matrix $SD$ and anomalous degree $AD$. We propose a local anomalous degree algorithm to speed up our method based on spectral clustering~\cite{spec_clu}. Spectral clustering is able to identify spatial communities of nodes in graph structures. According to several studies~\cite{tong2017simpler,yao2018deep,zheng2013time}, which assume that traffic in nearby locations should be similar, we also assume that flows in the same community and in the spatially nearby regions will be historically similar. Given a graph $\mathcal G$, we perform spectral decomposition and obtain $k$ graph spatial features of each flow. Then we use K-means~\cite{kmeans_KDD04}, a common unsupervised clustering method, to cluster flows into $k$ classes. 

\begin{algorithm}
	\caption{Local Anomalous Degree Algorithm}
	\label{alg:A}
	\begin{algorithmic}
		
		\REQUIRE Road graph $\mathcal G$
		\STATE 1. Compute the adjacency matrix $A$, degree matrix $D$, and normalized Laplacian matrix  $L = I-D^{-\frac{1}{2}}AD^{-\frac{1}{2}}$.
		\STATE 2. Compute the first $k$ eigenvectors $v_1, v_2, ..., v_C $ of $L$. 
		\STATE 3. Let $F \in \mathbb{R}^{N \times k}$ is the feature matrix of all nodes in the graph. \STATE
		\For{ node $i$ in $G$}
		{
			\STATE $F_i = \left[ v_{0,i}, v_{1,i}, ..., v_{k-1,i}\right] $
			
		}
		\STATE 4. Use K-means method to cluster nodes into $k$ classes ($k$ labels).  
		\STATE 5. Compute local-similarity matrix $S$ and local-similarity decrease matrix $SD$.  
		$S^{[t-T+1:t]}_{{\xi_i},{\xi_j}}=\left\{
		\begin{array}{rcl}
		0       &      & {, label_i \neq label_j}\\
		\\
		P\left(v^{[t-T+1:t]}_{\xi_i},v^{[t-T+1:t]}_{\xi_j}\right)   &      & {, label_i = label_j}\\
		\end{array} \right. $
		\STATE 6. Compute local-anomalous degree $AD$.  
		
		$	AD^t_{\xi_i} = \frac{\Sigma_{\xi_j \in HS^t_{\xi_i} \& (label_j = label_i)}S^{t-1}_{i,j} \cdot SD^{t}_{i,j} }{\Sigma_{\xi_j \in HS^t_{\xi_i}\&  (label_j = label_i)}S^{t-1}_{i,j} }$
	\end{algorithmic}
\end{algorithm}

\para{Validation of Local Algorithm.} Figure~\ref{fig:clu} shows the clustering result when $k=10$ (marked by different colors). The result shows that the eigenvectors can effectively capture spatial graph features. Our method divides New York City into 10 local districts which are conform to the real-world urban districts, e.g., the red area corresponds to the Manhattan area in New York City. Then we only need to compute the local values of $S$, $SD$ and $AD$ in the same district.
}

Next, different from anomaly detection, we aim at exploring the impact on traffic flows of different {\color{nb2}urban} traffic incidents. Also taking Figure~\ref{fig:example1:b} as an example, there is a flawed scene that three flows $\xi_1$, $\xi_{23}$ and $\xi_{385}$ are historically similar to each other at time slot $t$. Therefore, the {\color{nb}sharp variations} of $\xi_{23}$ and $\xi_{385}$ will strongly affect the anomalous degree of $\xi_{1}$. Figure~\ref{fig:me:a} shows the anomalous degrees of them from 4:00 am to 12:00 pm. Near 06:32 am, $\xi_1$ actually has a higher anomalous degree (0.198) than $\xi_{23}$ (0.110) and $\xi_{385}$ (0.085). However, we can see it intuitively in Figure~\ref{fig:example1:c} that when close to 06:32 am, the anomalous variation of speeds of $\xi_{23}$ and $\xi_{385}$ are more striking than $\xi_1$. The reason for this diametrically opposite result is that after the incident, the tendency of anomalous changes of $\xi_{23}$ and $\xi_{385}$ are mighty similar, which leads to the low anomalous degree of them. Therefore, in order to handle the scenario mentioned above, we add another metric to help amend our discovery method.

\begin{defn}{\bfseries{Relative Speed Variation (RSV). }}Given a flow $\xi_i$ at time slot $t$, and the historical speed sequence {\color{nb} $\left[v^{t-{T}+1}_{\xi_i},v^{t-{T}+2}_{\xi_i},\cdots ,v^{t}_{\xi_i} \right]$} of $\xi_i$ in a $T$-length time window, we define the relative speed variation of $\xi_i$ is

\begin{equation}
RSV^t_{\xi_i} = \left| \frac{\sum_{t'=t-{T}+1}^{t'=t}v_{\xi_i}^{t'}}{T}  - v_{\xi_i}^{t} \right| / max\left( v^{t_{s}}_{\xi_i},v^{t_{s}+1}_{\xi_i},\cdots,v^{t_{e}}_{\xi_i} \right)
\end{equation}

	We define a normalization time window and use the max value observed in the time window to normalize $RSV$. We use 24 hours (288 intervals) as the normalization window length, i.e., $t_s = t - 144 $ and $t_e = t + 144$, and $T=10$ intervals.
\end{defn}

{\color{nb2}
\para{Validation of $RSV$.} As a heuristic approach, we test different candidate computing methods of $RSV$ as baselines for validation. We consider three related features: slope of speed variation  ($k$) ~\cite{viovy1992best}, recent speed ($v^{t-1}$) and historical average speed ($ \bar v$)~\cite{boriboonsomsin2012eco} corresponding to three candidate computing methods of $RSV$. They are listed as follows:

\begin{itemize}
	\item[1)] Consider all three features: $RSV_{k+v^{t-1}+{\bar v}} = \left| {\bar v}-v^t \right| \times \bar{k} \times p + \left| v^{t-1} - v^t \right| \times k^{t-1} \times q $, where $p$ and $q$ is two parameters to control the ratio of recent speed and historical average speed, $\bar{k}$ is the historical average slope and $k^{t-1}$ is the slope of time slot $t-1$ and $t$.
	\item[2)] Consider recent speed and historical average speed: $RSV_{v^{t-1}+{\bar v}} = \left| {\bar v}-v^t \right| \times p + \left| v^{t-1} - v^t \right| \times q $.
	\item[3)] Consider historical average speed: $RSV_{{\bar v}} = \left| {\bar v}-v^t \right| $.
\end{itemize}
We use the normalized item to normalize the three computing methods. We use Pearson correlation coefficient to calculate the correlation coefficient of $AD$ and $RSV$ of all urban traffic incidents in our dataset (an hour before and after the incident). In order to use $RSV$ to amend $AD$, we choose the most negatively correlated computing method as our $RSV$ ($p$ and $q$ are set to 0.5), i.e., only consider historical average speed: $RSV_{{\bar v}} = \left| {\bar v}-v^t \right| $. Figure~\ref{fig:me:b} shows the $RSV$ result of the congestion incident. Near 06:32 am, in contrast to $AD$, the max ${RSV}$ of $\xi_{23}$ and $\xi_{385}$ are both larger (0.377 and 0.333) than $\xi_{1}$. It is conform to the speed variation (Figure~\ref{fig:example1:c}) and indicates that $RSV$ can also capture anomalies well and effectively correct the flaw of $AD$. 
}

\begin{defn}{\bfseries{Incident Effect Score (IES).}}  Due to the complementarity of anomalous degree $AD$ and relative speed variation $RSV$, we combine both of them to compute the incident effect score. Given a flow $\xi_i$ at time slot $t$, the incident effect score is calculated by: 
	\begin{equation}
		IES^t_{\xi_i}= \rho \, \cdot \, AD^t_{\xi_i} +  (1- \rho) \, \cdot \,RSV^t_{\xi_i},
	\end{equation}
	
		where $\rho$ is a parameter to control the ratio of $AD$ and $RSV$.
\end{defn}

\begin{defn}{\bfseries{Critical Incident}.} For incidents like mega-events, the traffic flows might be affected before incidents begin, on the contrary, incidents like traffic collisions will begin to affect traffic flows after they occurred. Therefore, given an incident $inci_k$ with a start time $t_{s}$, we firstly set a T-length ``start to influence'' window $W=\left[ t_s - \frac{T}{2}, t_{s}+\frac{T}{2} \right]$ and define the flows which are highly affected by the incident is 
	$\{\xi_i \mid max(IES^{t-\frac{T}{2}}_{\xi_i}, IES^{t-\frac{T}{2}+1}_{\xi_i}, \cdots, IES^{t+\frac{T}{2}}_{\xi_i}) \geq \theta \}$,
	 where $\theta$ is a threshold parameter. 
	 
	 When 
	 ${\left|\{\xi_i \mid max(IES^{t-\frac{T}{2}}_{\xi_i}, IES^{t-\frac{T}{2}+1}_{\xi_i},
	 	 \cdots, IES^{t+\frac{T}{2}}_{\xi_i}) \geq \theta \}  \right|}_{inci_k}  > 0 $, more specifically, there is at least one flow is highly affected by $inci_k$, we call $inci_k$ is a critical incident, where $\left| \cdot \right|$ denotes the cardinality of a set. We define an incident which is not a critical incident as a non-critical incident.

\end{defn}

\subsection{Evaluation and results}

\para{Parameter Setting.} The datasets we use here are listed in Table~\ref{tab:datasets}. We set $r=500\, m$ and one hour as the length of ``start to influence'' time window. 

\para{Varing $\rho$ and $\theta$.}
Figure~\ref{fig:inci_res} shows the number of critical incidents discovered when varying $\rho$ and $\theta$. In SFO, when $\theta=0$, most incidents are discovered as critical (1,706 out of 1,832 averagely), which indicates that most incidents indeed have an impact on traffic flows. There are a small number of incidents which almost have no impact (6.9\%, $\theta = 0$ and 12.2\%, $\theta = 0.05$), which further proves that treating all traffic incidents equally for traffic speed prediction is unreasonable. When $\theta$ rises ($\theta=$0.10, 0.15 or 0.20), there is a sharp reduction of critical incidents, which indicates the impact of incidents varies in degree. In order to discover incidents with high impact, we set $\rho=0.6$ and $\theta=0.15$ of SFO. The results of NYC is similar with SFO, most incidents are discovered as critical incident when $\theta$ is set to 0 or 0.05. Reductions also appear when $\theta$ rises. We set {\color{nb}$\rho=0.5$ and $\theta=0.10$} of NYC.

\begin{figure}[t]
		\centering
		\subfigure[Spatial of SFO]{
			\label{fig:incilo:sfo} 
			\includegraphics[width=2.5in,height=1.5in]{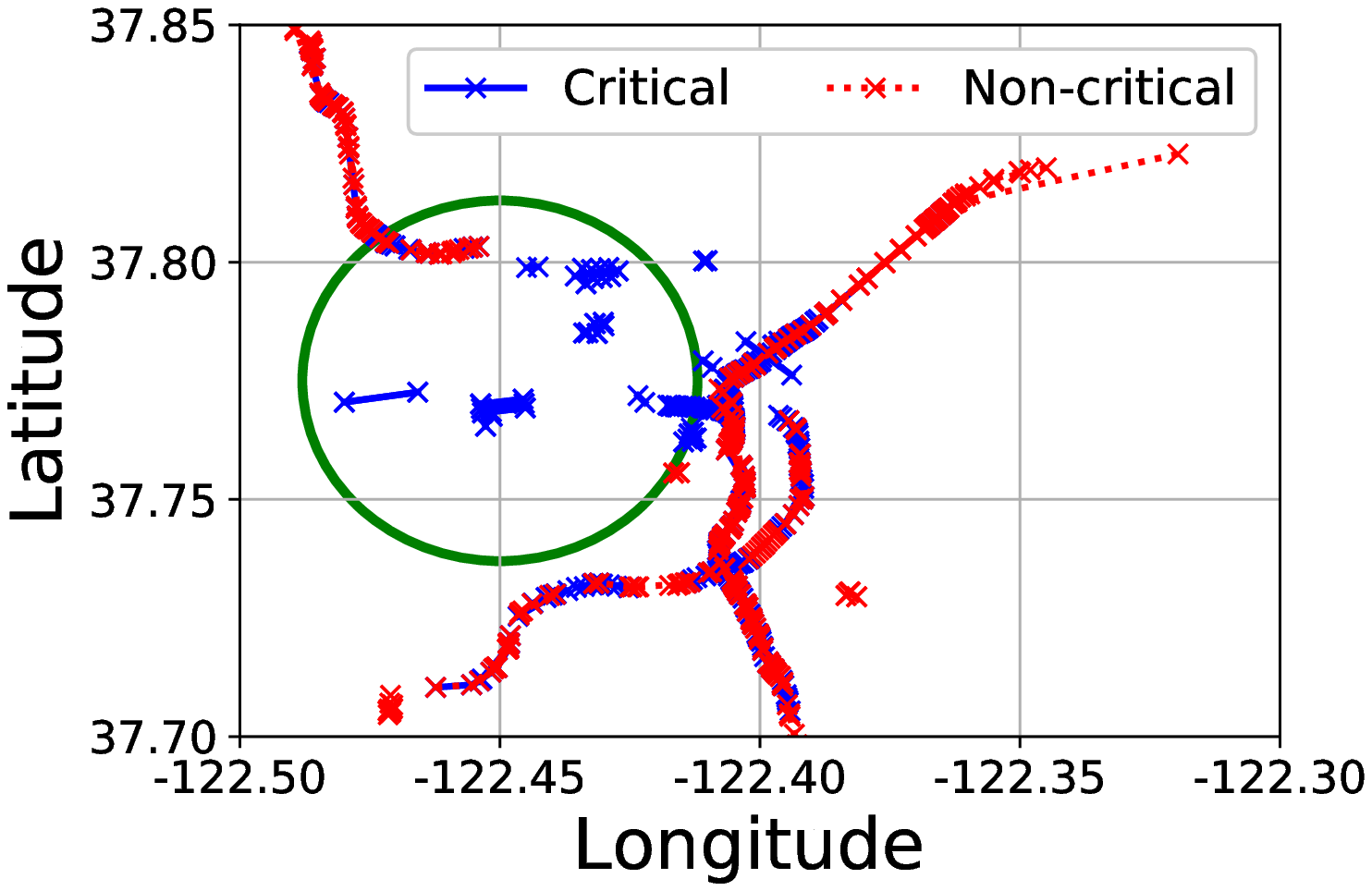}}
		\hspace{0.1in}
		\subfigure[Temporal of SFO]{
			\label{fig:incitime:sfo} 
			\includegraphics[width=2.5in,height=1.5in]{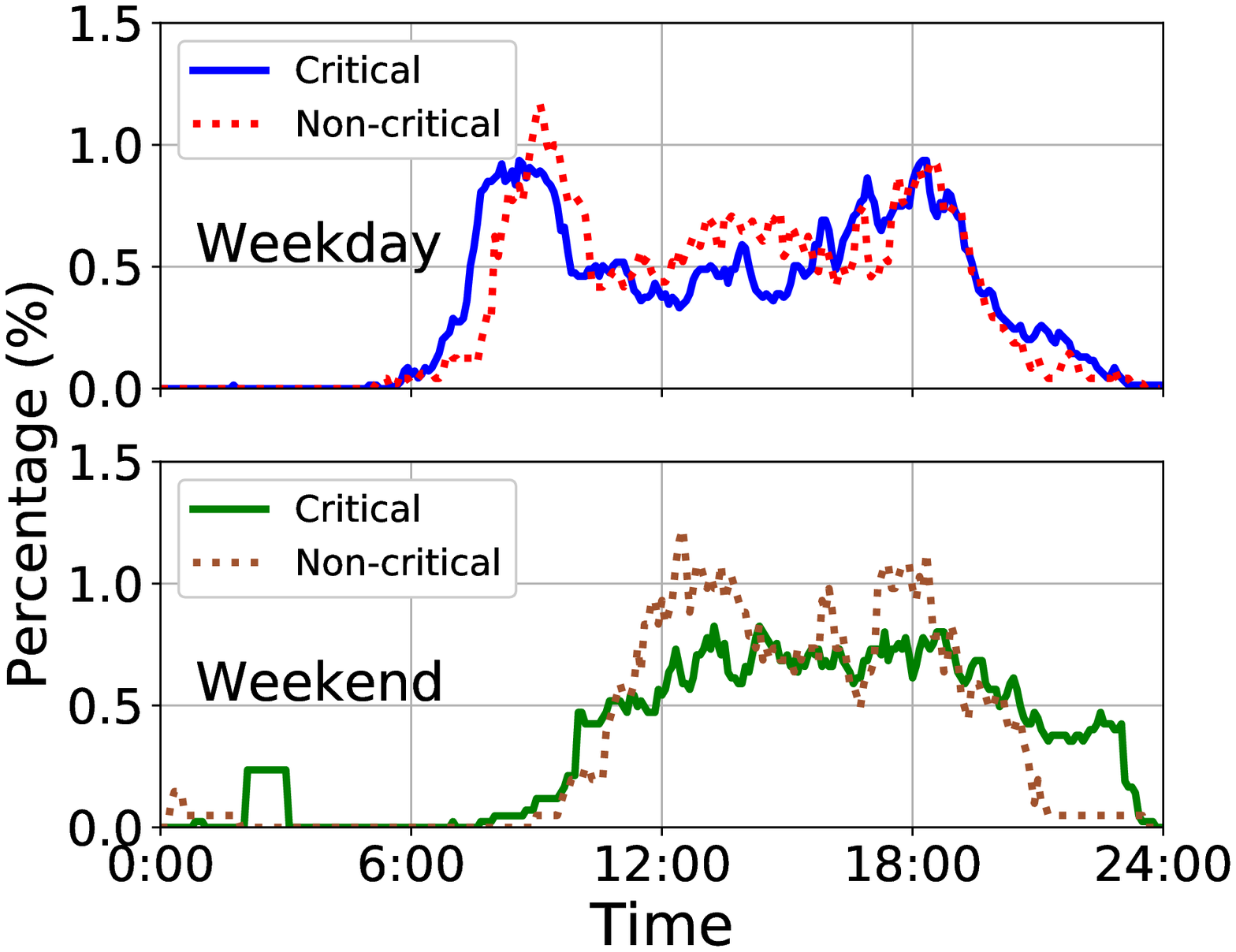}}
		\subfigure[Spatial of NYC]{
			\label{fig:incilo:nyc} 
			\includegraphics[width=2.5in,height=1.5in]{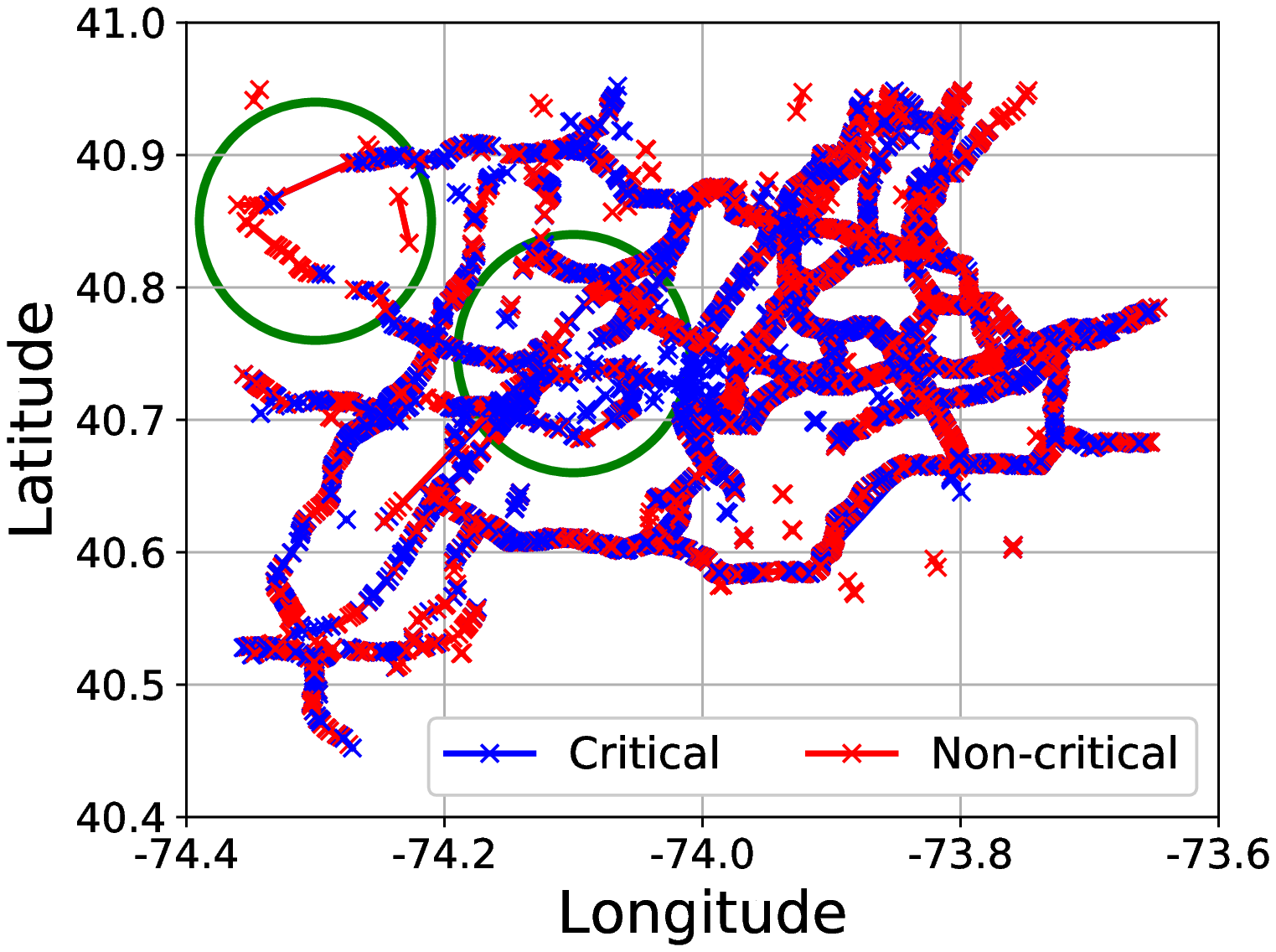}}
		\hspace{0.1in}
		\subfigure[Temporal of NYC]{
			\label{fig:incitime:nyc} 
			\includegraphics[width=2.5in,height=1.5in]{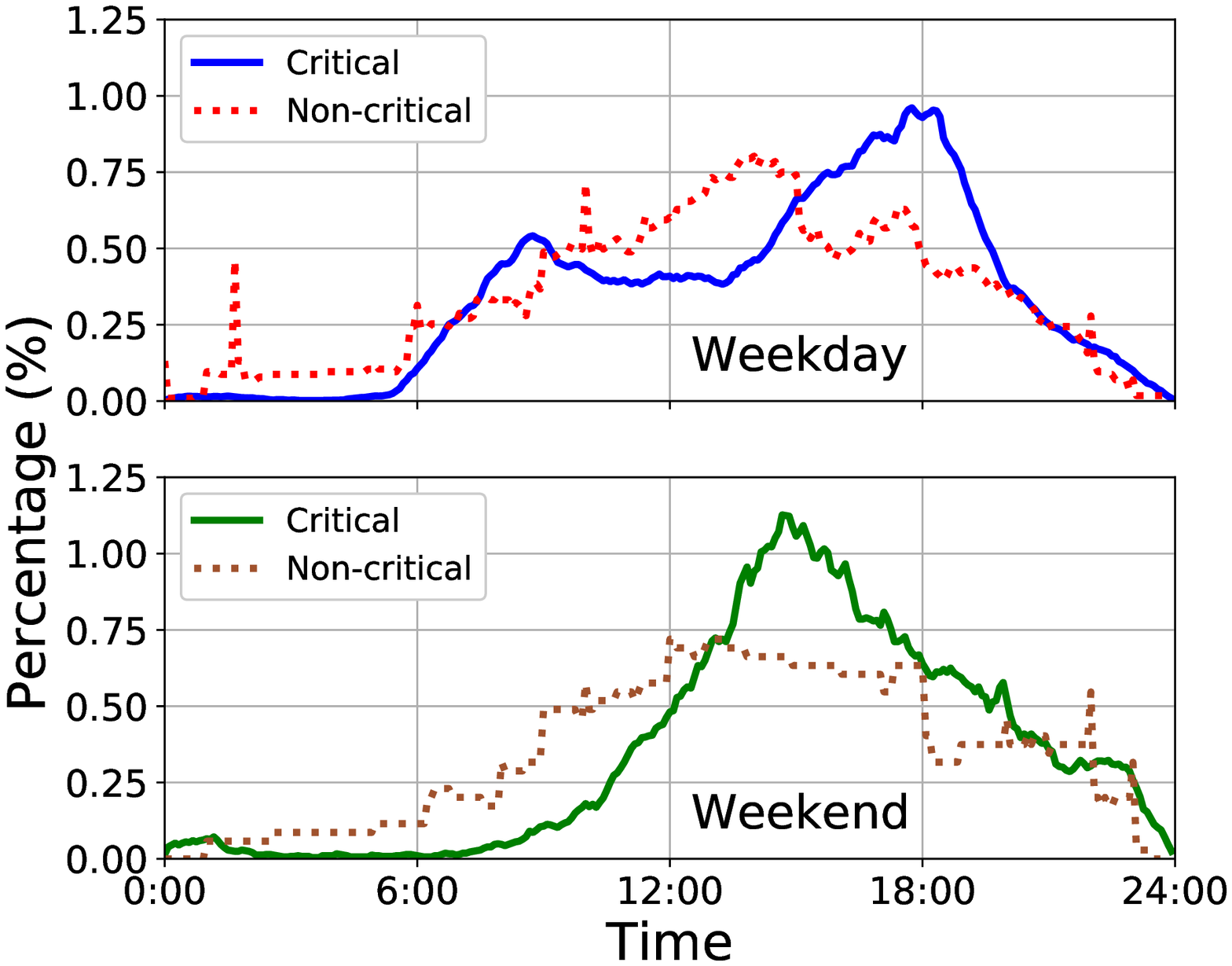}}

		\caption{{\color{nb2}Spatial and temporal distributions of urban traffic incidents}}
				\label{fig:inci:all} 
	
\end{figure}

\para{Spatial Distributions.} Figure~\ref{fig:incilo:sfo} and~\ref{fig:incilo:nyc} shows the spatial distributions of incidents in SFO and NYC. An incident is plotted by a line with an origin and an end. In SFO, although most of both two type incidents occur on the main roads (continuous parts), our method can effectively discover critical incidents (green circle). Moreover, we check critical incidents in the green circles and find they are mostly the Event type, which has a high severity level recorded by HERE. In NYC, both two type incidents also gather in the main roads. The number of urban traffic incidents in NYC is far more than in SFO but we can still observe the differences. Critical incidents which did not occur in the main roads are mainly locate in Manhattan (the middle circle). In the left circle, we find that most incidents away from city center are discovered as non-critical.

\para{Temporal Distributions.} Figure~\ref{fig:incitime:sfo} and~\ref{fig:incitime:nyc} show the temporal distributions of critical and non-critical incidents in two cities. In SFO, incidents mostly occur in rush hours (7-9 am and 4-7 pm), which is in line with daily routine. At about 12 pm (noon) and 3 pm on weekday, the ratio of critical incidents has a drop while the ratio of non-critical rises, which might because both time are not in rush hours and incidents may not have high impact. On weekend and during mid-afternoon, there is also a drop of the critical incidents and a rise of non-critical type. We also find that incidents are more likely to occur in the early morning on weekend than weekday. In NYC, most critical incidents also occur in rush hours. Incidents occur in the early morning tend to be non-critical in both two cities. On weekend, NYC only has one incident peak (mid-afternoon) and on weekday, NYC does not have the mid-afternoon peak while SFO presents the peak.

{\color{nb2}
\para{Summary of Results.} Parameters $\rho$ and $\theta$ represent the threshold to discover urban incidents with high impact on traffic speeds. The lower the $\theta$ and $\rho$ are, the lower the threshold to mark critical incidents. The results of varying $\rho$ and $\theta$ show that some urban incidents almost have no impact on traffic speeds and impact of urban incidents varies in degree, which indicate that it is unreasonable to use all urban traffic incidents features for traffic speed prediction. Spatio-temporal distributions show noteworthy differences between urban critical and non-critical incidents, which indicates that our urban critical incident discovery method can effectively discover incidents with high impact on traffic speeds.

}

\section{Extract the Latent Incident Impact Features}

\label{ignet}
So far, we have proven that our discovery method can effectively discover urban critical/non-critical incidents. In this section, we propose to use deep learning methods to extract the latent incident impact features for traffic speed prediction. Taking two aspects into account, we design a binary classifier to extract the latent impact features: 

\begin{itemize}
	\item There are some urban incidents almost have no impact on traffic flows and low-impact incidents features will even bring noise to the model. There are also noteworthy differences of spatio-temporal features between crucial and non-crucial incidents, which inspires us to consider the binary classification problem.
	\item The impact of urban incidents on traffic speeds varies in degree and the impact is neither binary nor strict multi-class. Therefore, we should not use the binary result directly, we propose to extract the latent impact features from the middle layer of the binary classifier for traffic speed prediction, where the latent features are continuous and filtered.
\end{itemize}

\subsection{Methodology}

\begin{figure}[t]
	\centering
	\includegraphics[width=5in,height=3.2in]{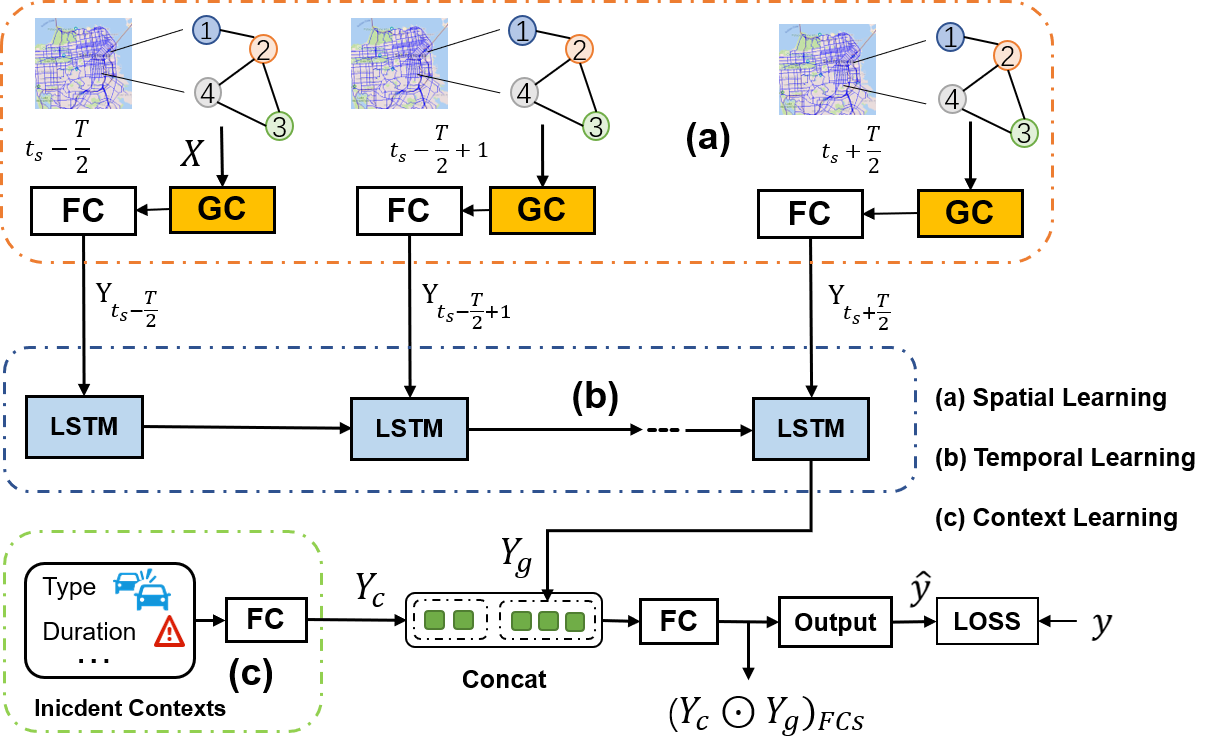}
	\caption{The architecture of the binary classifier}
	\label{fig:ignet} 
\end{figure}

\label{predict-incident}

The task of the binary classifier is to predict whether an incident is critical/non-critical, i.e., whether an urban incident has a high/low impact on traffic speed. Considering that the impact of incidents is related to spatio-temporal and context features and previous works~\cite{LCRNN_ijcai18, yao2018deep, DNN_SIG16} which use spatio-temporal and context features for traffic prediction (we will discuss them  in Section~\ref{related-work}), our classifier consists of three components: spatial learning component (GCN), temporal learning component (LSTM) and context learning component. 

\para{Spatial Learning: GCN (Figure~\ref{fig:ignet}(a)).}
City road network has latent traffic patterns and there are complex spatial dependencies~\cite{DCRNN_ICLR18}. We need to capture the road topological features, i.e., the spatial dependencies of the road network. Traditional methods divide the city into several grids and use Convolutional Neural Network (CNN) to capture spatial features~\cite{yao2018deep, DNN_SIG16}. However, it neglects the road topological features and also lose the spatial information within grids. Moreover, graph structure related features are hard to be used in CNN for our problem. We adopt graph convolutional network (GCN)~\cite{GCN_ICLR14} to learn the spatial topology features. GCN is known for being able to capture the topology features in non-Euclidean structures, which is suitable for road network. GCN model $f\left(X,A\right)$ follows the layer-wise propagation rule~\cite{GC_ICLR17}: 
\begin{equation} \label{gcn_1}
H^{l+1}=\sigma\left( \tilde{D}^{-\frac{1}{2}} \tilde{A} \tilde{D}^{-\frac{1}{2}} H^{\left(l\right)}W^{\left(l\right)}\right),
\end{equation}

	{\color{nb2}where $A$ is the adjacency matrix, $\tilde{A}$ is the adjacency matrix of the graph with added self-connections, $D$ is the degree matrix and ${\tilde{D}}_{ii}  = \Sigma_j{\tilde{A}_{ij}} $.} $L=\tilde{D}^{-\frac{1}{2}} \tilde{A} \tilde{D}^{-\frac{1}{2}}$ is the normalized Laplacian matrix of the graph $\mathcal G$. $\sigma$ denotes an activation function. $W$ is the trainable weight matrix, $H^{\left(l\right)} \in \mathbb{R}^{N \times D} $ is the matrix of activations in the $l$-th layer. $H^{\left(0\right)}=X$, where $X$ is the input vectors of GCN. 

We use the above mentioned graph $\mathcal G$. At each time slot $t$, we obtain a real-time speed of every flow in $\mathcal G$, and we define the speed snapshot $G^t = \{V^t_{\xi_0}, V^t_{\xi_1}, ..., V^t_{\xi_N-1} \}$, where $N$ is the total number of flows in the city. We also add another graph structure related feature: the distance of each flow from the incident, which is because of the impact of incidents on flows has a strong correlation with distance~\cite{tong2017simpler,yao2018deep,zheng2013time}. We define the distance $D_{\xi_i}$ of $\xi_i$ from the incident is the Euclidean distance between the flow center and incident center. Therefore, at each time slot $t$, the input features $X=\left[\left( V^t_{\xi_0}, D_{\xi_0} \right), \left( V^t_{\xi_1}, D_{\xi_1} \right), ..., \left( V^t_{\xi_N-1}, D_{\xi_N-1} \right)\right]$. For a urban traffic incident, the time span of input speed snapshots is $\left[ t_s - \frac{T}{2}, t_{s}+\frac{T}{2} \right]$, where $t_s$ is the start time of the incident and $T$ is the length of ``start to influence'' time window which is defined in Section~\ref{sec:incident}. 

For the input signal $X \in \mathbb{R}^{N \times C}$ with C input channels ($C=2$ here) {\color{nb2}and $F$ filters or features of spectral convolutions map as follows~\cite{GC_ICLR17}}:
	\begin{equation}\label{z}
	Z=\tilde{D}^{-\frac{1}{2}} \tilde{A} \tilde{D}^{-\frac{1}{2}}X\Theta
	, 
	\end{equation}
{\color{nb2}where $\Theta \in \mathbb{R}^{C \times F}$ is a matrix of filter parameters, $Z \in \mathbb{R}^{N \times F}$ is the convolved signal matrix and $F$ is the number of filters or features.} 
Next, at each time slot $t$, after k graph convolutional (GC) layers, we then feed middle states $H^k_t$ into $m$ fully connected (FC) layers to get the spatial learning output  $Y_t$ of each snapshot.

\para{Temporal Learning: LSTM (Figure~\ref{fig:ignet}(b)).} We feed a sequence of graph speed snapshots to GCN, and the output is a sequence of spatial features {\color{nb}at} each time slot from $t_s-\frac{T}{2}$ to $t_s+\frac{T}{2}$. Then we adopt Long Short-Term Memory (LSTM) model~\cite{LSTM} as our temporal learning component. LSTM is known for being able to learn long-term dependency information of time related sequences. LSTM has the ability to remove or add information to the state of the cell through a well-designed structure ``gate''. we extract the spatial features $Y_t$ for each snapshot in GCN and feed the sequence $\left[ Y_{t_s-\frac{T}{2}}, Y_{t_s-\frac{T}{2}+1}, \cdots, Y_{t_s+\frac{T}{2}} \right]$ into LSTM cells. Then we can iteratively get the output sequence $\left[ h_{t_s-\frac{T}{2}},h_{t_s-\frac{T}{2}+1}, \cdots, h_{t_s+\frac{T}{2}} \right]$. We use the last LSTM cell output as the output $Y_g$ of temporal learning part. 

\para{Context Learning (Figure~\ref{fig:ignet}(c)).} Incident context features are also important for prediction. We use the following features for context learning:

\begin{itemize}
	\item Incident type (e.g., traffic collision and event).
	\item Road status: Whether the urban incident lead to a road close or not.
	\item Start and end hour: HERE gives a start time $t_s$ and an anticipative end time $t_e$ of an incident.
	\item Incident duration: The anticipative duration of the incident, i.e., $t_e - t_s$.
	\item Weekday, Saturday or Sunday.
\end{itemize}

We use one-hot encoding to preprocess class features and normalize the incident duration feature. {\color{nb2}The context learning component is a Deep Neural Network (DNN) structure, more specifically, an input layer and a fully connected layer (shown in Figure~\ref{fig:ignet}(c)).} After embedding the context information, we feed the context embedding to a fully connected layer to get $Y_c$, which is the output of context learning.

\para{Latent incident impact features extraction.} After getting $Y_c$ and spatio-temporal feature $Y_g$, we use a concat operation to concatenate them as $Y_c \odot Y_g$ of each incident. Then we feed $Y_c \odot Y_g$ to $m$ FC layers. We extract the output of the last FC layer before the output layer as the latent incident impact features, which is because that output layer uses these features as the input to predict whether the incident has high impact on traffic flows. We denote the latent impact features as $(Y_c \odot Y_g)_{FCs}$. Finally we get the prediction value $\hat{y}$, and compute the loss compare to real value $y$.

\para{Objective Function and Evaluation Metric.} The classifier is training by minimizing Binary Cross Entropy Loss (BCELoss) between the predicted speed and the real value. BCELoss is defined as follow:
\begin{equation}
BCELoss=-\left(y \cdot log\left(\hat{y}\right) + \left(1-y\right)\cdot log\left(1-\hat{y}\right)  \right)
\end{equation}
We use BCELoss and F1-score ($F1-score= \frac{2 \cdot precision \cdot recall}{precision + recall}$) to evaluate the binary classifier. 

\subsection{Middle Experiments}

\para{Parameter Setting.} The datasets we use here are listed in Table~\ref{tab:datasets}. We use the discovery results obtained in {\color{nb}the} last section as the ground truth. There are 1,061 positive samples (critical) and 771 negative samples (non-critical) of SFO and 17,924 positive samples and 15367 negative samples of NYC. We use 5 minutes as the time interval and train our classifier with the following hyper-parameter settings: learning rate (0.001) with Adam optimizer. In GCN, we set two GCN layers followed by one FC layer with the 64-dimension output. The length of "start to influence" window is set to one hour, i.e., the input size of the first GCN layer is 12. We use \emph{relu} activation function and add {\color{nb}Dropout ($d=0.8$)} in GC layer.  We use one LSTM layer with {\color{nb}64-dimension} hidden states. After concat, we adopt one FC layer (16-dimension) and follow by the output FC layer using \emph{sigmoid} activation function. We use 70\% data for training, and the remaining 30\% as the test set. We select 90\% of training set for training and 10\% as the validation set for early stopping.

\para{Results and Analysis.} Using the traffic incident and traffic speed sub-datasets for training, we finally get 0.8241 F1-score and 0.4429 BCELoss in the test set of SFO, and  4731 BCELoss and 0.8000 F1-score of NYC. Our binary classifier model can capture the latent impact features on traffic flows of different incidents, more specifically, we can get the embedding $Y_c \odot Y_g$ of each input incident. $Y_c$ is the output features of context learning and $Y_g$ is the output features of spatio-temporal learning. And we feed $Y_c \odot Y_g$ into $m$ ($m=1$ in our experiment) FC layers to extract the latent impact features $(Y_c \odot Y_g)_{FCs}$ before the ouput layer. We will use the binary classifier in the next section as an internal component to help improve traffic speed prediction performance. {\color{nb2}Since we take the classifier as a middleware of our incident-driven framework, we further evaluate our complete framework with competitive baselines in the next Section.}

\section{Incident-driven Traffic Speed Prediction}
\label{experiments}
So far, we can effectively capture the latent impact features of urban incidents on traffic flow speeds. Combining above methods, we propose Deep Incident-Aware Graph Convolutional Network (DIGC-Net) to improve traffic speed prediction by urban incident data.

\begin{figure}[t]
	\centering
	\includegraphics[width=5.5in,height=3.5in]{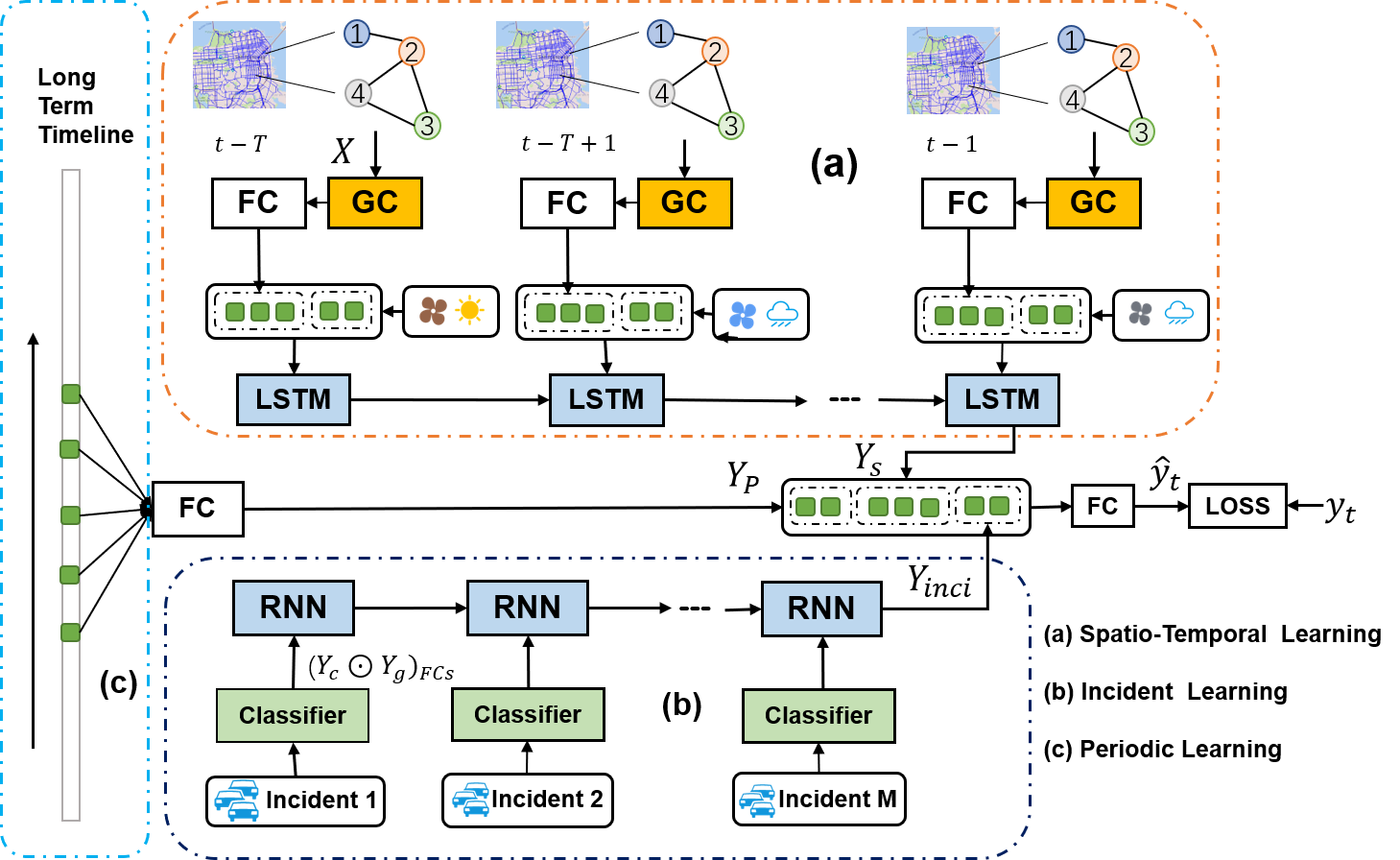}
	\caption{The architecture of DIGC-Net}
	\label{fig:final} 
\end{figure}

\subsection{Methodology}
DIGC-Net (Figure~\ref{fig:final}) consists of three components: spatio-temporal, incident and periodic learning. Our prediction problem is defined above in the Section~\ref{overview}.

\para{Spatio-temporal Learning (Figure~\ref{fig:final}(a)).} {\color{nb2}Considering traffic speed prediction also related to spatio-temporal patterns of traffic network and previous works~\cite{LCRNN_ijcai18, yao2018deep, DNN_SIG16} which use spatio-temporal features for traffic prediction (we will discuss them in Section~\ref{related-work}), we use the similar structure of spatial and temporal learning in the binary classifier.}  The spatial-temporal and context structure is a common use in traffic prediction, and we use GCN rather than CNN to better capture spatial features of road network here. GCN is used for capturing spatial graph features and LSTM is adopted to capture the time evolution patterns of traffic speeds. The input features of each node is $ V^t_{\xi_i} $ in GCN, i.e., the speed of each flow at time slot $t$. More specifically, the input features is $X^t=\left[V^t_{\xi_0}, V^t_{\xi_1}, ..., V^t_{\xi_{N-1}}\right]$, which is graph speed snapshot at time slot $t$. We input a sequence of graph speed snapshots features $[X^{t-T}, X^{t-T+1},X^{t-1}]$ to GCN and after the GCN part, similar to~\cite{yao2018deep}, we concatenate the weather contexts at each time slot $t$ to get $Y_{t}$. Then we feed the spatial features sequence $\left[ Y_{t-T}, Y_{t-T+1}, \cdots, Y_{t-1} \right]$ to LSTM cells to iteratively get the output sequence $\left[ h_{t-T},h_{t-T+1}, \cdots, h_{t-1} \right]$. Then we use $k$ learnable units to predict $k$ future traffic speeds $\left[Y_S^{t}, Y_S^{t+1}, \cdots, Y_S^{t+k-1}\right]$. The output of spatio-temporal learning is $Y_S$. 

\para{Incident Learning (Figure~\ref{fig:final}(b)).} To predict traffic speed at time slot $t$, {\color{nb}we select all incidents occurred within $[t-125 min, t-5 min]$ as the incident learning inputs (the last two hours), where $t-125min$ is the earliest included incident occurrence time and $t-5min$ is the latest time}. We use the pre-trained binary classifier (trained in last Section) to extract $(Y_c \odot Y_g)_{FCs}$, i.e., the latent incident impact features of each incident. Because the number of incidents occur within the time range is uncertain and incidents occur in a sequential order, so we adopt standard Recurrent Neural Network (RNN)~\cite{mikolov2010recurrent} for incident learning. RNN is a neural network that contains loops that allow information to be persisted. {\color{nb2} Previous incidents will affect the traffic conditions, which may lead to the occurrence of future incidents. Using RNN also help us capture the interrelation of sequentially occurring urban traffic incidents, which is neglected by previous works~\cite{lin2017road}}. We denote $Y_{inci}$ as the output of the last RNN cell.

\para{Periodic Learning (Figure~\ref{fig:final}(c)).} Traffic flow speeds change periodically and we use the similar structure of~\cite{LCRNN_ijcai18} to learn long-term periodical patterns. We use the same time slots in the last 5 days to learn the periodic features. A fully connected layer is adopted to capture the long-term periodic features. The output of periodic learning is $Y_{P}$.

\para{Output.}
After getting spatio-temporal features $Y_S$, incident impact features $Y_{inci}$, and periodic features $Y_P$, we adopt a concat operation to concatenate them, then feed them into $m$ FC-layers. Finally we get the prediction value $\hat{y_t}$, and {\color{nb}compute the loss compare to the real value $y_t$}.

{\color{nb2}
\para{Objective Function and Evaluation Metric.} DIGC-net is training by minimizing Mean Squared Error ($MSE= \sum_{i=1}^{N}{(\hat{y_i}-{y_i})}^2$)  between the predicted speed and the real value. We use Mean Absolute Percentage Error to evaluate DIGC-Net, MAPE is defined as follow:
\begin{equation}
MAPE=\frac{100\%}{N}\sum_{i=1}^{N} |\frac{\hat{y_i}-{y_i}}{y_i}| ,
\end{equation}

	where $N$ is the total number of flows.
}

\begin{table}[t]%
	\centering
	\caption{{\color{nb}Evaluation among different methods}}
	\begin{tabular}{|p{2.2 cm}  | l| l|}
		\toprule
		Method             &       MAPE-SFO   & MAPE-NYC \\
		\midrule[1pt]
		ARIMA     &  26.70 \% & 38.60 \%    \\
		\midrule[1pt]
		SVR       &  28.24 \%   &  39.73 \% \\
		\midrule[1pt]
		LSTM      &  18.98 \% & 30.26 \%\\
		\midrule[1pt]
		GC     & 15.69 \%  & 25.79 \% \\
		\midrule[1pt]
		LSM-RN      & 13.72 \%   & 21.53 \%  \\
		\bottomrule
		LC-RNN      &  12.26 \%  & 18.77 \% \\
		\bottomrule
		\bfseries{DIGC-Net}   &  \bfseries{11.02} \%  & \bfseries{17.21} \% \\
		\bottomrule
	\end{tabular}
	\label{tab:results}
\end{table}%

\subsection{Evaluations}

\para{Parameter Setting.} The datasets we use here are listed in Table~\ref{tab:datasets}. {\color{nb} We set 5 minutes as the time interval and time window as 4 hours, i.e., $T=48$}.  We train our network {\color{nb}with the following hyper-parameter settings}: learning rate (0.001) with Adam optimizer. In spatio-temporal learning, we set two GCN layers followed by one FC-layer (64-dimension) and the input size of the first GCN layer is 64. We use \emph{relu} activation function and add Dropout in GCN layer with $d=0.5$.  In incident learning, we use one RNN layer with 128-dimension hidden state. In periodical learning, we use one FC layer with 64-dimension hidden state. After {\color{nb}concat operation}, we adopt one FC-layer with 256-dimension and connect the final output layer. We use \emph{relu} activation function in the FC layers. We use first three weeks data for training, and the remaining one week data as the test set. In training dataset, we select 90\% of them for training and 10\% as the validation set for early stopping.

{\color{nb2}
\para{Comparison with competitive benchmarks.} We compare our model with the following models in consideration of covering traditional machine learning, matrix decomposition and state-of-the-art deep learning methods: 

\begin{itemize}
	\item[1)] ARIMA~\cite{arima}: Autoregressive integrated moving average is a classics linear model in time series forecasting.
	\item[2)] SVR~\cite{svr}: Support Vector Regression is based on the computation of linear regression in a high dimensional feature space and is widely used.
	\item[3)] LSTM~\cite{ma2015long}: This method uses LSTM to capture non-linear traffic dynamic to predict traffic speed.
	\item[4)] GC~\cite{GC_NIPS16}: GC uses graph convolution, pooling and fully-connected layer to predict future traffic speed. GC is the variation of basic GCN with the efficient pooling. 
	\item[5)] LSM-RN~\cite{LSMRN_kdd16}: Latent space model for road networks learns the attributes of vertices in latent spaces which mainly uses matrix decomposition. It also consider spatio-temporal effects of latent attributes and use an incremental online algorithm to predict traffic speed.
	\item[6)] LC-RNN~\cite{LCRNN_ijcai18}: LC-RNN takes advantage of both RNN and CNN models and designs a Look-up operation to capture complex traffic evolution patterns, which outperforms ST-ResNet~\cite{zhang2017deep} and DCNN~\cite{ma2017learning}, so we do not further compare ST-ResNet and DCNN here.
\end{itemize}

Table~\ref{tab:results} shows the MAPE results of using different methods of SFO and NYC. All other benchmarks in the table is one-step prediction. When compared with different methods, DIGC-Net achieves the best performance in both two cities. DIGC-net has relatively from 10.11\% up to 60.97\% lower MAPE than these benchmarks in SFO and relatively from 8.31\% up to 56.68\% lower MAPE than these benchmarks in NYC. We also note significant variance between SFO and NYC among all methods, likely due to large differences in the traffic road network (NYC is much larger than SFO: 2,416 vs 13,028 nodes and 19,334 vs 92,470 edges). The results indicate that DIGC-net can effectively incorporate incident, spatio-temporal, periodic and context features for traffic speed prediction.

\para{Comparison with variants of DIGC-net.} We also present the comparison with different variants of DIGC-net with only spatio-temporal component, spatio-temporal and periodic component, and the whole DIGC-net with all components (spatio-temporal, periodic and incident component). The results are shown in Table~\ref{tab:variants}. The first finding is that the performance improvement of periodic learning is relatively weak, with only difference of 0.25\% of SFO and 0.06\% of NYC. One possible reason that the improvement margin of SFO is larger than NYC is that there is a relatively simple road network in SFO and the variation of traffic speed is more regular. The MAPE without incident learning (spatio-temporal + periodic) is 12.22\% of SFO and 18.63 \% of NYC, which also outperform all benchmarks (sightly outperform LC-RNN). It also verifies that our incident learning component is the key to the improvement with a 1.2\% MAPE improvement of SFO and 1.42\% MAPE improvement of NYC.

\begin{table}[h]%
	\centering
	\caption{{\color{nb}Evaluation among different variants of DIGC-net}}
	\begin{tabular}{|p{9 cm}  | l| l|}
		\toprule
		Variant             &       MAPE-SFO   & MAPE-NYC \\
		\midrule[1pt]
		Spatio-temporal    &  12.47 \% & 18.69 \%    \\
		\midrule[1pt]
		Spatio-temporal + periodic     &  12.22 \%   &  18.63 \% \\
		\midrule[1pt]
		\bfseries{DIGC-Net-all}  \bfseries{ (Spatio-temporal + periodic + incident )}   & \bfseries{11.02} \%  & \bfseries{17.21} \% \\
		\bottomrule
	\end{tabular}
	\label{tab:variants}
\end{table}%

\para{Comparison with different time period.}
As shown in Figure~\ref{fig:incitime:sfo} and Figure~\ref{fig:incitime:nyc},  the number of incidents varies over time, and more incidents occur at traffic peak periods. Meanwhile, traffic speed variation is also time-sensitive. Therefore, we further select 2:00 - 3:00 am as the wee hour and 07:00 - 08:00 am as the rush hour, and take SFO as the illustration to evaluate the performance of different methods. Figure~\ref{fig:res:all} shows the MAPE results in the wee hour and rush hour. In the wee hour, our method has relatively from 2.08\% up to 64.43\% lower MAPE than these benchmarks in SFO, and relatively from 10.78\% up to 89.50\% lower MAPE than these benchmarks in the rush hour. The performance of our method and LC-RNN are pretty similar in the wee hour but exhibits a relatively clear gap in the rush hour, which derives from more complex traffic patterns in the rush hour.

\begin{figure}[h]
	\begin{minipage}[t]{1.0\columnwidth}
		\centering
		\subfigure[Wee Hour]{
			\label{fig:res:sf_wee} 
			\includegraphics[width=2.5in,height=1.8in]{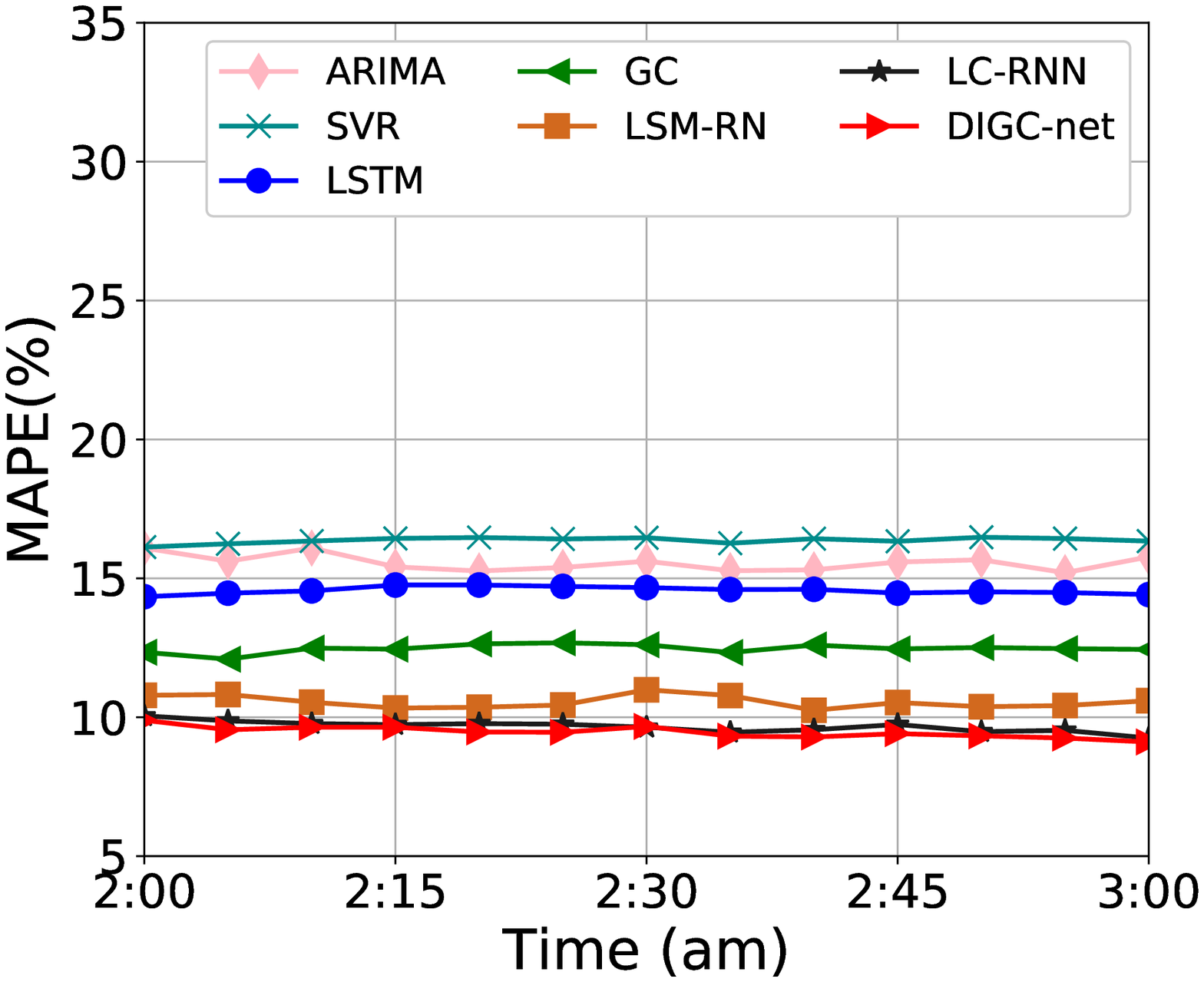}}
		\hspace{0.25in}
		\subfigure[Rush Hour]{
			\label{fig:res:sf_rush} 
			\includegraphics[width=2.5in,height=1.8in]{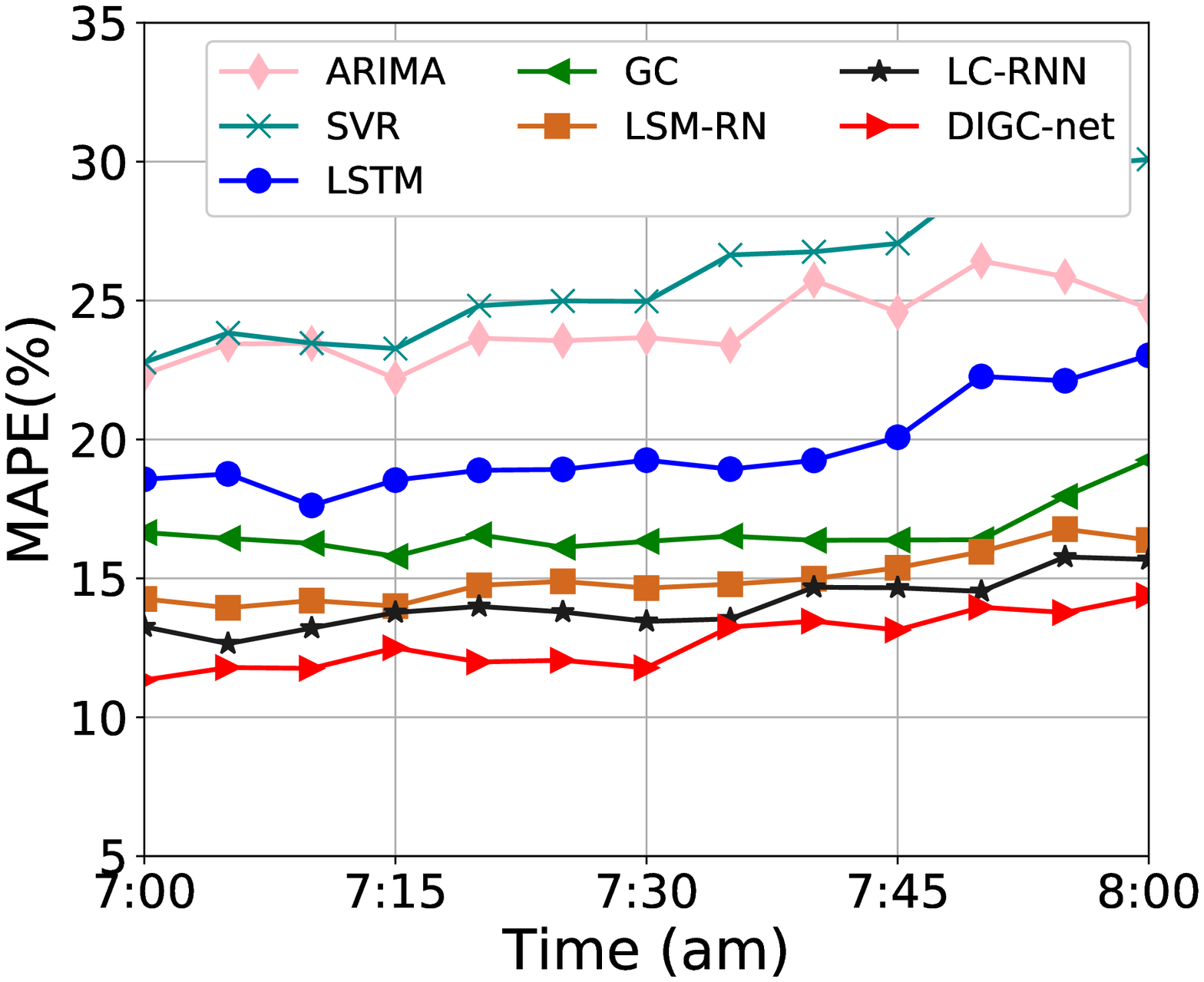}}
		
		\caption{{\color{nb}Time-sensitive comparison of SFO}}
		\label{fig:res:all} 
		
	\end{minipage}
	\vspace{-0.1in}
\end{figure}

\para{Comparison for multi-step prediction.} We then present the comparison results for multi-step prediction. DIGC-net can be used for multi-step speed prediction by setting $k$ learnable units in spatio-temporal learning component. We set prediction length $k=1, 2, 3$ (speeds of next 5, 10 and 15 minutes) to evaluate the multi-step prediction case. The results are shown in Table~\ref{tab:multi-step}. The performance of DIGC-net of multi-step prediction remains stable as the predicted length increases (drop relatively 3.09\% of $k=2$ and 5.44 \% of $k=3$ compare to $k=1$ in SFO and drop relatively 3.88\% of $k=2$ and 9.03\% of $k=3$ compare to $k=1$ in NYC). When prediction length is within three steps, DIGC-net outperforms all other baselines of one-step prediction in SFO, and in NYC, only of one-step that LC-RNN outperforms three-steps DIGC-net. The multi-step results demonstrate that our model can be effectively applied to multi-step prediction within a certain time range.

\begin{table}[h]%
	\centering
	\caption{Evaluation for multi-step prediction}
	\begin{tabular}{|p{2.5 cm}  | l| l|}
		\toprule
		Method             &       MAPE-SFO   & MAPE-NYC \\
		\bottomrule
		\bfseries{DIGC-Net, k=1}    &  \bfseries{11.02} \%  & \bfseries{17.27} \% \\
		\bottomrule
		DIGC-Net, k=2    &  11.36 \%  & 17.94 \% \\
		\bottomrule
		DIGC-Net, k=3    &   11.62 \% & 18.83 \% \\
		\bottomrule
	\end{tabular}
	\label{tab:multi-step}
\end{table}%

}

\section{Related Work}
\label{related-work}

\para{Traffic Speed Prediction.}
A number of solutions have been proposed for traffic speed prediction. ARIMA~\cite{arima} is a classical model for this area, and regression methods~\cite{castro2009online} are also widely used for predicting traffic speed. There are also matrix spectral decomposition models for traffic speed prediction: \cite{LSMRN_kdd16} proposed a latent space model to capture both topological and temporal properties. Recently, deep learning approachs achieve great success in this space by using spatio-temporal and context features~\cite{lv2014traffic,ma2017learning}. {\color{nb2} The spatio-temporal and context structure is a common use in traffic prediction.} \cite{DNN_SIG16} divided road network into grids and used CNN to capture spatial dependencies. \cite{LCRNN_ijcai18} proposed a model that integrates both RNN and CNN models. GCN begin to be used for traffic speed prediction recently because of the ability to effectively capture the topology features in non-Euclidean structures. \cite{DCRNN_ICLR18} proposed to model the traffic flow as a diffusion process on a directed graph. \cite{yu2017spatio} proposed the STGCN model to tackle the time series prediction problem in traffic domain. In our work, we effectively incorporate traffic incident, spatio-temporal, periodic and weather features for traffic speed prediction. Our main contributions are focus on the effective utilization of incident information for improving prediction performance.

\para{Urban Incidents.} Research on urban anomalous incidents mainly focus on the detection of incidents.  \cite{incident_social} mined tweet texts to extract incident information to do the traffic incident detection. \cite{Detect_Anomalies_ubicomp18} proposed an algorithm based on SVM to capture rare patterns to detect urban anomalies. \cite{convlstm_KDD18} proposed a ConvLSTM model for traffic incident prediction. There are also a few works focus on mining the impact of incidents. \cite{INCI_URB12} proposed a system for predicting the cost and impact of highway incidents, in order to classify the duration of the incident induced delays and the magnitude of the incident impact. \cite{javid2018framework} developed a framework to estimate travel time variability caused by traffic incidents by using a series of robust regression methods. In our work, we extract the latent incident impact features for traffic speed prediction.

\section{Conclusion}
\label{conclusion}
In this work, we investigate the problem of incident-driven traffic speed prediction. We first propose the critical incident discovery method to identify urban crucial incidents and their impact on traffic flows. Then we design a binary classifier to extract the latent incident impact features for improving traffic speed prediction. Combining both processes, we propose a Deep Incident-Aware Graph Convolutional Network (DIGC-Net) to effectively incorporate traffic incident, spatio-temporal, periodic and weather features
for traffic speed prediction. We evaluate DIGC-Net using two real-world urban traffic datasets of large cities (SFO and NYC). The results demonstrate the superior performance of DIGC-Net and validate the effectiveness of extracting latent incident features in our framework.

\bibliographystyle{ACM-Reference-Format}
\bibliography{myref}


\begin{thebibliography}{42}


\ifx \showCODEN    \undefined \def \showCODEN     #1{\unskip}     \fi
\ifx \showDOI      \undefined \def \showDOI       #1{#1}\fi
\ifx \showISBNx    \undefined \def \showISBNx     #1{\unskip}     \fi
\ifx \showISBNxiii \undefined \def \showISBNxiii  #1{\unskip}     \fi
\ifx \showISSN     \undefined \def \showISSN      #1{\unskip}     \fi
\ifx \showLCCN     \undefined \def \showLCCN      #1{\unskip}     \fi
\ifx \shownote     \undefined \def \shownote      #1{#1}          \fi
\ifx \showarticletitle \undefined \def \showarticletitle #1{#1}   \fi
\ifx \showURL      \undefined \def \showURL       {\relax}        \fi
\providecommand\bibfield[2]{#2}
\providecommand\bibinfo[2]{#2}
\providecommand\natexlab[1]{#1}
\providecommand\showeprint[2][]{arXiv:#2}

\bibitem[\protect\citeauthoryear{Boriboonsomsin, Barth, Zhu, and
  Vu}{Boriboonsomsin et~al\mbox{.}}{2012}]%
        {boriboonsomsin2012eco}
\bibfield{author}{\bibinfo{person}{Kanok Boriboonsomsin},
  \bibinfo{person}{Matthew~J Barth}, \bibinfo{person}{Weihua Zhu}, {and}
  \bibinfo{person}{Alexander Vu}.} \bibinfo{year}{2012}\natexlab{}.
\newblock \showarticletitle{Eco-routing navigation system based on multisource
  historical and real-time traffic information}.
\newblock \bibinfo{journal}{\emph{IEEE Transactions on Intelligent
  Transportation Systems}} \bibinfo{volume}{13}, \bibinfo{number}{4}
  (\bibinfo{year}{2012}), \bibinfo{pages}{1694--1704}.
\newblock


\bibitem[\protect\citeauthoryear{Bruna, Zaremba, Szlam, and LeCun}{Bruna
  et~al\mbox{.}}{2014}]%
        {GCN_ICLR14}
\bibfield{author}{\bibinfo{person}{Joan Bruna}, \bibinfo{person}{Wojciech
  Zaremba}, \bibinfo{person}{Arthur Szlam}, {and} \bibinfo{person}{Yann
  LeCun}.} \bibinfo{year}{2014}\natexlab{}.
\newblock \showarticletitle{Spectral networks and locally connected networks on
  graphs}. In \bibinfo{booktitle}{\emph{Proceedings of 2nd International
  Conference on Learning Representations}} \emph{(\bibinfo{series}{ICLR '14})}.
\newblock


\bibitem[\protect\citeauthoryear{Castro-Neto, Jeong, Jeong, and
  Han}{Castro-Neto et~al\mbox{.}}{2009}]%
        {castro2009online}
\bibfield{author}{\bibinfo{person}{Manoel Castro-Neto},
  \bibinfo{person}{Young-Seon Jeong}, \bibinfo{person}{Myong-Kee Jeong}, {and}
  \bibinfo{person}{Lee~D Han}.} \bibinfo{year}{2009}\natexlab{}.
\newblock \showarticletitle{Online-SVR for short-term traffic flow prediction
  under typical and atypical traffic conditions}.
\newblock \bibinfo{journal}{\emph{Expert systems with applications}}
  \bibinfo{volume}{36}, \bibinfo{number}{3} (\bibinfo{year}{2009}),
  \bibinfo{pages}{6164--6173}.
\newblock


\bibitem[\protect\citeauthoryear{Contreras, Espinola, Nogales, and
  Conejo}{Contreras et~al\mbox{.}}{2003}]%
        {arima}
\bibfield{author}{\bibinfo{person}{Javier Contreras}, \bibinfo{person}{Rosario
  Espinola}, \bibinfo{person}{Francisco~J Nogales}, {and}
  \bibinfo{person}{Antonio~J Conejo}.} \bibinfo{year}{2003}\natexlab{}.
\newblock \showarticletitle{ARIMA models to predict next-day electricity
  prices}.
\newblock \bibinfo{journal}{\emph{IEEE transactions on power systems}}
  \bibinfo{volume}{18}, \bibinfo{number}{3} (\bibinfo{year}{2003}),
  \bibinfo{pages}{1014--1020}.
\newblock


\bibitem[\protect\citeauthoryear{Deng, Shahabi, Demiryurek, Zhu, Yu, and
  Liu}{Deng et~al\mbox{.}}{2016}]%
        {LSMRN_kdd16}
\bibfield{author}{\bibinfo{person}{Dingxiong Deng}, \bibinfo{person}{Cyrus
  Shahabi}, \bibinfo{person}{Ugur Demiryurek}, \bibinfo{person}{Linhong Zhu},
  \bibinfo{person}{Rose Yu}, {and} \bibinfo{person}{Yan Liu}.}
  \bibinfo{year}{2016}\natexlab{}.
\newblock \showarticletitle{Latent Space Model for Road Networks to Predict
  Time-Varying Traffic}. In \bibinfo{booktitle}{\emph{Proceedings of the 22nd
  ACM SIGKDD International Conference on Knowledge Discovery and Data Mining}}
  \emph{(\bibinfo{series}{KDD '16})}. \bibinfo{pages}{1525--1534}.
\newblock


\bibitem[\protect\citeauthoryear{Dhillon, Guan, and Kulis}{Dhillon
  et~al\mbox{.}}{2004}]%
        {kmeans_KDD04}
\bibfield{author}{\bibinfo{person}{Inderjit~S. Dhillon},
  \bibinfo{person}{Yuqiang Guan}, {and} \bibinfo{person}{Brian Kulis}.}
  \bibinfo{year}{2004}\natexlab{}.
\newblock \showarticletitle{Kernel K-means: Spectral Clustering and Normalized
  Cuts}. In \bibinfo{booktitle}{\emph{Proceedings of the 10th ACM SIGKDD
  international conference on Knowledge Discovery and Data Dining}}
  \emph{(\bibinfo{series}{KDD '04})}. \bibinfo{pages}{551--556}.
\newblock


\bibitem[\protect\citeauthoryear{Gao, Guo, Sun, Dai, Zhu, Hu, and Li}{Gao
  et~al\mbox{.}}{2019}]%
        {gao2019aggressive}
\bibfield{author}{\bibinfo{person}{Ruipeng Gao}, \bibinfo{person}{Xiaoyu Guo},
  \bibinfo{person}{Fuyong Sun}, \bibinfo{person}{Lin Dai},
  \bibinfo{person}{Jiayan Zhu}, \bibinfo{person}{Chenxi Hu}, {and}
  \bibinfo{person}{Haibo Li}.} \bibinfo{year}{2019}\natexlab{}.
\newblock \showarticletitle{Aggressive driving saves more time? multi-task
  learning for customized travel time estimation}. In
  \bibinfo{booktitle}{\emph{Proceedings of the 28th International Joint
  Conference on Artificial Intelligence (IJCAI '19)}}. AAAI Press,
  \bibinfo{pages}{1689--1696}.
\newblock


\bibitem[\protect\citeauthoryear{Gu, Qian, and Chen}{Gu et~al\mbox{.}}{2016}]%
        {incident_social}
\bibfield{author}{\bibinfo{person}{Yiming Gu}, \bibinfo{person}{Zhen(Sean)
  Qian}, {and} \bibinfo{person}{Feng Chen}.} \bibinfo{year}{2016}\natexlab{}.
\newblock \showarticletitle{From Twitter to detector: Real-time traffic
  incident detection using social media data}.
\newblock \bibinfo{journal}{\emph{Transportation {R}esearch {P}art {C}:
  {E}merging {T}echnologies.}}  \bibinfo{volume}{67} (\bibinfo{year}{2016}),
  \bibinfo{pages}{321--342}.
\newblock


\bibitem[\protect\citeauthoryear{He, Rong, Liu, and Du}{He
  et~al\mbox{.}}{2019}]%
        {he2019traffic}
\bibfield{author}{\bibinfo{person}{Ya-qin He}, \bibinfo{person}{Yu-lun Rong},
  \bibinfo{person}{Zu-peng Liu}, {and} \bibinfo{person}{Sheng-pin Du}.}
  \bibinfo{year}{2019}\natexlab{}.
\newblock \showarticletitle{Traffic Influence Degree of Urban Traffic Accident
  Based on Speed Ratio}.
\newblock \bibinfo{journal}{\emph{Journal of Highway and Transportation
  Research and Development (English Edition)}} \bibinfo{volume}{13},
  \bibinfo{number}{3} (\bibinfo{year}{2019}), \bibinfo{pages}{96--102}.
\newblock


\bibitem[\protect\citeauthoryear{Here}{Here}{2019}]%
        {Here}
\bibfield{author}{\bibinfo{person}{Here}.} \bibinfo{year}{2019}\natexlab{}.
\newblock \bibinfo{booktitle}{\emph{https://developer.here.com/.}}
\newblock


\bibitem[\protect\citeauthoryear{Hochreiter and Schmidhuber}{Hochreiter and
  Schmidhuber}{1997}]%
        {LSTM}
\bibfield{author}{\bibinfo{person}{Sepp Hochreiter} {and}
  \bibinfo{person}{Jurgen Schmidhuber}.} \bibinfo{year}{1997}\natexlab{}.
\newblock \showarticletitle{Long Short-term Memory}.
\newblock \bibinfo{journal}{\emph{Neural computation.}} (\bibinfo{year}{1997}),
  \bibinfo{pages}{1735--1780}.
\newblock


\bibitem[\protect\citeauthoryear{Javid and Javid}{Javid and Javid}{2018}]%
        {javid2018framework}
\bibfield{author}{\bibinfo{person}{Roxana~J Javid} {and}
  \bibinfo{person}{Ramina~Jahanbakhsh Javid}.} \bibinfo{year}{2018}\natexlab{}.
\newblock \showarticletitle{A framework for travel time variability analysis
  using urban traffic incident data}.
\newblock \bibinfo{journal}{\emph{IATSS research}} \bibinfo{volume}{42},
  \bibinfo{number}{1} (\bibinfo{year}{2018}), \bibinfo{pages}{30--38}.
\newblock


\bibitem[\protect\citeauthoryear{Johnson, Henderson, Perry, Sch{\"o}ning, and
  Hecht}{Johnson et~al\mbox{.}}{2017}]%
        {johnson2017beautiful}
\bibfield{author}{\bibinfo{person}{Isaac Johnson}, \bibinfo{person}{Jessica
  Henderson}, \bibinfo{person}{Caitlin Perry}, \bibinfo{person}{Johannes
  Sch{\"o}ning}, {and} \bibinfo{person}{Brent Hecht}.}
  \bibinfo{year}{2017}\natexlab{}.
\newblock \showarticletitle{Beautiful… but at What Cost?: An Examination of
  Externalities in Geographic Vehicle Routing}.
\newblock \bibinfo{journal}{\emph{Proceedings of the ACM on Interactive,
  Mobile, Wearable and Ubiquitous Technologies (Ubicomp '17)}}
  \bibinfo{volume}{1}, \bibinfo{number}{2} (\bibinfo{year}{2017}),
  \bibinfo{pages}{15}.
\newblock


\bibitem[\protect\citeauthoryear{Kipf and Welling}{Kipf and Welling}{2017}]%
        {GC_ICLR17}
\bibfield{author}{\bibinfo{person}{Thomas~N. Kipf} {and} \bibinfo{person}{Max
  Welling}.} \bibinfo{year}{2017}\natexlab{}.
\newblock \showarticletitle{Semi-Supervised Classification with Graph
  Convolutional Networks}. In \bibinfo{booktitle}{\emph{Proceedings of 5th
  International Conference on Learning Representations}}
  \emph{(\bibinfo{series}{ICLR '17})}.
\newblock


\bibitem[\protect\citeauthoryear{Li, Yu, Shahabi, and Liu}{Li
  et~al\mbox{.}}{2018}]%
        {DCRNN_ICLR18}
\bibfield{author}{\bibinfo{person}{Yaguang Li}, \bibinfo{person}{Rose Yu},
  \bibinfo{person}{Cyrus Shahabi}, {and} \bibinfo{person}{Yan Liu}.}
  \bibinfo{year}{2018}\natexlab{}.
\newblock \showarticletitle{DIFFUSION CONVOLUTIONAL RECURRENT NEURAL NETWORK:
  DATA-DRIVEN TRAFFIC FORECASTING}. In \bibinfo{booktitle}{\emph{Proceedings of
  6th International Conference on Learning Representations.}}
  \emph{(\bibinfo{series}{ICLR '18})}.
\newblock


\bibitem[\protect\citeauthoryear{Li, Liu, Xu, Duan, and Wang}{Li
  et~al\mbox{.}}{2017}]%
        {li2017reinforcement}
\bibfield{author}{\bibinfo{person}{Zhibin Li}, \bibinfo{person}{Pan Liu},
  \bibinfo{person}{Chengcheng Xu}, \bibinfo{person}{Hui Duan}, {and}
  \bibinfo{person}{Wei Wang}.} \bibinfo{year}{2017}\natexlab{}.
\newblock \showarticletitle{Reinforcement learning-based variable speed limit
  control strategy to reduce traffic congestion at freeway recurrent
  bottlenecks}.
\newblock \bibinfo{journal}{\emph{IEEE transactions on intelligent
  transportation systems}} \bibinfo{volume}{18}, \bibinfo{number}{11}
  (\bibinfo{year}{2017}), \bibinfo{pages}{3204--3217}.
\newblock


\bibitem[\protect\citeauthoryear{Lin, Li, Chen, Ye, and Huai}{Lin
  et~al\mbox{.}}{2017}]%
        {lin2017road}
\bibfield{author}{\bibinfo{person}{Lu Lin}, \bibinfo{person}{Jianxin Li},
  \bibinfo{person}{Feng Chen}, \bibinfo{person}{Jieping Ye}, {and}
  \bibinfo{person}{Jinpeng Huai}.} \bibinfo{year}{2017}\natexlab{}.
\newblock \showarticletitle{Road traffic speed prediction: a probabilistic
  model fusing multi-source data}.
\newblock \bibinfo{journal}{\emph{IEEE Transactions on Knowledge and Data
  Engineering}} \bibinfo{volume}{30}, \bibinfo{number}{7}
  (\bibinfo{year}{2017}), \bibinfo{pages}{1310--1323}.
\newblock


\bibitem[\protect\citeauthoryear{Lin}{Lin}{1989}]%
        {Pearson}
\bibfield{author}{\bibinfo{person}{Lawrence I-Kuei Lin}.}
  \bibinfo{year}{1989}\natexlab{}.
\newblock \showarticletitle{A Concordance Correlation Coefficient to Evaluate
  Reproducibility}.
\newblock \bibinfo{journal}{\emph{Biometrics.}}  \bibinfo{volume}{67}
  (\bibinfo{year}{1989}), \bibinfo{pages}{255--268}.
\newblock


\bibitem[\protect\citeauthoryear{Lv, Duan, Kang, Li, and Wang}{Lv
  et~al\mbox{.}}{2014}]%
        {lv2014traffic}
\bibfield{author}{\bibinfo{person}{Yisheng Lv}, \bibinfo{person}{Yanjie Duan},
  \bibinfo{person}{Wenwen Kang}, \bibinfo{person}{Zhengxi Li}, {and}
  \bibinfo{person}{Fei-Yue Wang}.} \bibinfo{year}{2014}\natexlab{}.
\newblock \showarticletitle{Traffic flow prediction with big data: a deep
  learning approach}.
\newblock \bibinfo{journal}{\emph{IEEE Transactions on Intelligent
  Transportation Systems}} \bibinfo{volume}{16}, \bibinfo{number}{2}
  (\bibinfo{year}{2014}), \bibinfo{pages}{865--873}.
\newblock


\bibitem[\protect\citeauthoryear{Lv, Xu, Zheng, Yin, Zhao, and Zhou}{Lv
  et~al\mbox{.}}{2018}]%
        {LCRNN_ijcai18}
\bibfield{author}{\bibinfo{person}{Zhongjian Lv}, \bibinfo{person}{Jiajie Xu},
  \bibinfo{person}{Kai Zheng}, \bibinfo{person}{Hongzhi Yin},
  \bibinfo{person}{Pengpeng Zhao}, {and} \bibinfo{person}{Xiaofang Zhou}.}
  \bibinfo{year}{2018}\natexlab{}.
\newblock \showarticletitle{LC-RNN: A Deep Learning Model for Traffic Speed
  Prediction}. In \bibinfo{booktitle}{\emph{Proceedings of the Twenty-Seventh
  International Joint Conference on Artificial Intelligence}}
  \emph{(\bibinfo{series}{IJCAI '18})}.
\newblock


\bibitem[\protect\citeauthoryear{Ma, Dai, He, Ma, Wang, and Wang}{Ma
  et~al\mbox{.}}{2017}]%
        {ma2017learning}
\bibfield{author}{\bibinfo{person}{Xiaolei Ma}, \bibinfo{person}{Zhuang Dai},
  \bibinfo{person}{Zhengbing He}, \bibinfo{person}{Jihui Ma},
  \bibinfo{person}{Yong Wang}, {and} \bibinfo{person}{Yunpeng Wang}.}
  \bibinfo{year}{2017}\natexlab{}.
\newblock \showarticletitle{Learning traffic as images: a deep convolutional
  neural network for large-scale transportation network speed prediction}.
\newblock \bibinfo{journal}{\emph{Sensors}} \bibinfo{volume}{17},
  \bibinfo{number}{4} (\bibinfo{year}{2017}), \bibinfo{pages}{818}.
\newblock


\bibitem[\protect\citeauthoryear{Ma, Tao, Wang, Yu, and Wang}{Ma
  et~al\mbox{.}}{2015}]%
        {ma2015long}
\bibfield{author}{\bibinfo{person}{Xiaolei Ma}, \bibinfo{person}{Zhimin Tao},
  \bibinfo{person}{Yinhai Wang}, \bibinfo{person}{Haiyang Yu}, {and}
  \bibinfo{person}{Yunpeng Wang}.} \bibinfo{year}{2015}\natexlab{}.
\newblock \showarticletitle{Long short-term memory neural network for traffic
  speed prediction using remote microwave sensor data}.
\newblock \bibinfo{journal}{\emph{Transportation Research Part C: Emerging
  Technologies}}  \bibinfo{volume}{54} (\bibinfo{year}{2015}),
  \bibinfo{pages}{187--197}.
\newblock


\bibitem[\protect\citeauthoryear{Michaël, Bresson, and Vandergheynst}{Michaël
  et~al\mbox{.}}{2016}]%
        {GC_NIPS16}
\bibfield{author}{\bibinfo{person}{Defferrard Michaël},
  \bibinfo{person}{Xavier Bresson}, {and} \bibinfo{person}{Pierre
  Vandergheynst}.} \bibinfo{year}{2016}\natexlab{}.
\newblock \showarticletitle{Convolutional neural networks on graphs with fast
  localized spectral filtering}. In \bibinfo{booktitle}{\emph{Proceedings of
  Neural Information Processing Systems.}} \emph{(\bibinfo{series}{NIPS '16})}.
\newblock


\bibitem[\protect\citeauthoryear{Mikolov, Karafi{\'a}t, Burget,
  {\v{C}}ernock{\`y}, and Khudanpur}{Mikolov et~al\mbox{.}}{2010}]%
        {mikolov2010recurrent}
\bibfield{author}{\bibinfo{person}{Tom{\'a}{\v{s}} Mikolov},
  \bibinfo{person}{Martin Karafi{\'a}t}, \bibinfo{person}{Luk{\'a}{\v{s}}
  Burget}, \bibinfo{person}{Jan {\v{C}}ernock{\`y}}, {and}
  \bibinfo{person}{Sanjeev Khudanpur}.} \bibinfo{year}{2010}\natexlab{}.
\newblock \showarticletitle{Recurrent neural network based language model}. In
  \bibinfo{booktitle}{\emph{Eleventh annual conference of the international
  speech communication association}}.
\newblock


\bibitem[\protect\citeauthoryear{Miller and Gupta}{Miller and Gupta}{2012}]%
        {INCI_URB12}
\bibfield{author}{\bibinfo{person}{Mahalia Miller} {and}
  \bibinfo{person}{Chetan Gupta}.} \bibinfo{year}{2012}\natexlab{}.
\newblock \showarticletitle{Mining Traffic Incidents to Forecast Impact}. In
  \bibinfo{booktitle}{\emph{Proceedings of the ACM SIGKDD International
  Workshop on Urban Computing}} \emph{(\bibinfo{series}{Urbcomp '12})}.
\newblock


\bibitem[\protect\citeauthoryear{Pan, Demiryurek, and Shahabi}{Pan
  et~al\mbox{.}}{2012}]%
        {RUSH_ICDM12}
\bibfield{author}{\bibinfo{person}{Bei Pan}, \bibinfo{person}{Ugur Demiryurek},
  {and} \bibinfo{person}{Cyrus Shahabi}.} \bibinfo{year}{2012}\natexlab{}.
\newblock \showarticletitle{Utilizing Real-World Transportation Data for
  Accurate Traffic Prediction}. In \bibinfo{booktitle}{\emph{Proceedings of
  2012 IEEE 12th International Conference on Data Mining}}
  \emph{(\bibinfo{series}{ICDM '18})}.
\newblock


\bibitem[\protect\citeauthoryear{Rathore, Ahmad, Paul, and Rhob}{Rathore
  et~al\mbox{.}}{2016}]%
        {rathore2016urban}
\bibfield{author}{\bibinfo{person}{M~Mazhar Rathore}, \bibinfo{person}{Awais
  Ahmad}, \bibinfo{person}{Anand Paul}, {and} \bibinfo{person}{Seungmin Rhob}.}
  \bibinfo{year}{2016}\natexlab{}.
\newblock \showarticletitle{Urban planning and building smart cities based on
  the internet of things using big data analytics}.
\newblock \bibinfo{journal}{\emph{Computer Networks}}  \bibinfo{volume}{101}
  (\bibinfo{year}{2016}), \bibinfo{pages}{63--80}.
\newblock


\bibitem[\protect\citeauthoryear{Rumsey}{Rumsey}{2015}]%
        {rumsey2015u}
\bibfield{author}{\bibinfo{person}{Deborah~J Rumsey}.}
  \bibinfo{year}{2015}\natexlab{}.
\newblock \showarticletitle{U Can: statistics for dummies}.
\newblock  (\bibinfo{year}{2015}).
\newblock


\bibitem[\protect\citeauthoryear{Smola and Schölkopf}{Smola and
  Schölkopf}{1997}]%
        {svr}
\bibfield{author}{\bibinfo{person}{Alex~J. Smola} {and}
  \bibinfo{person}{Bernhard Schölkopf}.} \bibinfo{year}{1997}\natexlab{}.
\newblock \showarticletitle{A Tutorial on Support Vector Regression}.
\newblock \bibinfo{journal}{\emph{Statistics and Computing.}}
  (\bibinfo{year}{1997}), \bibinfo{pages}{199--222}.
\newblock


\bibitem[\protect\citeauthoryear{Tong, Chen, Zhou, Chen, Wang, Yang, Ye, and
  Lv}{Tong et~al\mbox{.}}{2017}]%
        {tong2017simpler}
\bibfield{author}{\bibinfo{person}{Yongxin Tong}, \bibinfo{person}{Yuqiang
  Chen}, \bibinfo{person}{Zimu Zhou}, \bibinfo{person}{Lei Chen},
  \bibinfo{person}{Jie Wang}, \bibinfo{person}{Qiang Yang},
  \bibinfo{person}{Jieping Ye}, {and} \bibinfo{person}{Weifeng Lv}.}
  \bibinfo{year}{2017}\natexlab{}.
\newblock \showarticletitle{The simpler the better: a unified approach to
  predicting original taxi demands based on large-scale online platforms}. In
  \bibinfo{booktitle}{\emph{Proceedings of the 23rd ACM SIGKDD international
  conference on knowledge discovery and data mining}}. ACM,
  \bibinfo{pages}{1653--1662}.
\newblock


\bibitem[\protect\citeauthoryear{Viovy, Arino, and Belward}{Viovy
  et~al\mbox{.}}{1992}]%
        {viovy1992best}
\bibfield{author}{\bibinfo{person}{N Viovy}, \bibinfo{person}{O Arino}, {and}
  \bibinfo{person}{AS Belward}.} \bibinfo{year}{1992}\natexlab{}.
\newblock \showarticletitle{The Best Index Slope Extraction (BISE): A method
  for reducing noise in NDVI time-series}.
\newblock \bibinfo{journal}{\emph{International Journal of remote sensing}}
  \bibinfo{volume}{13}, \bibinfo{number}{8} (\bibinfo{year}{1992}),
  \bibinfo{pages}{1585--1590}.
\newblock


\bibitem[\protect\citeauthoryear{Xie, Chen, Xiao, and Wang}{Xie
  et~al\mbox{.}}{2018}]%
        {City_Behavior_IEEE18}
\bibfield{author}{\bibinfo{person}{Rong Xie}, \bibinfo{person}{Yang Chen},
  \bibinfo{person}{Yu Xiao}, {and} \bibinfo{person}{Xin Wang}.}
  \bibinfo{year}{2018}\natexlab{}.
\newblock \showarticletitle{We Know Your Preferences in New Cities: Mining and
  Modeling the Behavior of Travelers}.
\newblock \bibinfo{journal}{\emph{IEEE Communications Magazine}}
  (\bibinfo{year}{2018}), \bibinfo{pages}{pages 28--35}.
\newblock


\bibitem[\protect\citeauthoryear{Yahoo}{Yahoo}{2019}]%
        {Yahoo_Weather}
\bibfield{author}{\bibinfo{person}{Yahoo}.} \bibinfo{year}{2019}\natexlab{}.
\newblock \bibinfo{booktitle}{\emph{https://developer.yahoo.com/weather/.}}
\newblock


\bibitem[\protect\citeauthoryear{Yao, Tang, Wei, Zheng, and Li}{Yao
  et~al\mbox{.}}{2019}]%
        {yao2019revisiting}
\bibfield{author}{\bibinfo{person}{Huaxiu Yao}, \bibinfo{person}{Xianfeng
  Tang}, \bibinfo{person}{Hua Wei}, \bibinfo{person}{Guanjie Zheng}, {and}
  \bibinfo{person}{Zhenhui Li}.} \bibinfo{year}{2019}\natexlab{}.
\newblock \showarticletitle{Revisiting spatial-temporal similarity: A deep
  learning framework for traffic prediction}. In \bibinfo{booktitle}{\emph{AAAI
  Conference on Artificial Intelligence (AAAI '19)}}.
\newblock


\bibitem[\protect\citeauthoryear{Yao, Wu, Ke, Tang, Jia, Lu, Gong, Ye, and
  Li}{Yao et~al\mbox{.}}{2018}]%
        {yao2018deep}
\bibfield{author}{\bibinfo{person}{Huaxiu Yao}, \bibinfo{person}{Fei Wu},
  \bibinfo{person}{Jintao Ke}, \bibinfo{person}{Xianfeng Tang},
  \bibinfo{person}{Yitian Jia}, \bibinfo{person}{Siyu Lu},
  \bibinfo{person}{Pinghua Gong}, \bibinfo{person}{Jieping Ye}, {and}
  \bibinfo{person}{Zhenhui Li}.} \bibinfo{year}{2018}\natexlab{}.
\newblock \showarticletitle{Deep multi-view spatial-temporal network for taxi
  demand prediction}. In \bibinfo{booktitle}{\emph{Thirty-Second AAAI
  Conference on Artificial Intelligence (AAAI' 18)}}.
\newblock


\bibitem[\protect\citeauthoryear{Yu, Yin, and Zhu}{Yu et~al\mbox{.}}{2017}]%
        {yu2017spatio}
\bibfield{author}{\bibinfo{person}{Bing Yu}, \bibinfo{person}{Haoteng Yin},
  {and} \bibinfo{person}{Zhanxing Zhu}.} \bibinfo{year}{2017}\natexlab{}.
\newblock \showarticletitle{Spatio-temporal graph convolutional networks: A
  deep learning framework for traffic forecasting}.
\newblock \bibinfo{journal}{\emph{arXiv preprint arXiv:1709.04875}}
  (\bibinfo{year}{2017}).
\newblock


\bibitem[\protect\citeauthoryear{Yu and Shi}{Yu and Shi}{2003}]%
        {spec_clu}
\bibfield{author}{\bibinfo{person}{Stella~X. Yu} {and} \bibinfo{person}{Jianbo
  Shi}.} \bibinfo{year}{2003}\natexlab{}.
\newblock \showarticletitle{Multiclass spectral clustering}. In
  \bibinfo{booktitle}{\emph{Proceedings Ninth IEEE International Conference on
  Computer Vision}} \emph{(\bibinfo{series}{ICCV '03})}.
\newblock


\bibitem[\protect\citeauthoryear{Yuan, Zhou, and Yang}{Yuan
  et~al\mbox{.}}{2018}]%
        {convlstm_KDD18}
\bibfield{author}{\bibinfo{person}{Zhuoning Yuan}, \bibinfo{person}{Xun Zhou},
  {and} \bibinfo{person}{Tianbao Yang}.} \bibinfo{year}{2018}\natexlab{}.
\newblock \showarticletitle{Hetero-ConvLSTM: A deep learning approach to
  traffic accident prediction on heterogeneous spatio-temporal data}. In
  \bibinfo{booktitle}{\emph{Proceedings of the 24th ACM SIGKDD International
  Conference on Knowledge Discovery \& Data Mining}}
  \emph{(\bibinfo{series}{KDD '18})}. \bibinfo{pages}{984--992}.
\newblock


\bibitem[\protect\citeauthoryear{Zhang, Zheng, and Yu}{Zhang
  et~al\mbox{.}}{2018}]%
        {Detect_Anomalies_ubicomp18}
\bibfield{author}{\bibinfo{person}{Huichu Zhang}, \bibinfo{person}{Yu Zheng},
  {and} \bibinfo{person}{Yong Yu}.} \bibinfo{year}{2018}\natexlab{}.
\newblock \showarticletitle{Detecting urban anomalies using multiple
  Spatio-temporal data sources}.
\newblock \bibinfo{journal}{\emph{Proceedings of the ACM on Interactive,
  Mobile, Wearable and Ubiquitous Technologies (Ubicomp '18)}}
  \bibinfo{volume}{2}, \bibinfo{number}{1} (\bibinfo{year}{2018}),
  \bibinfo{pages}{54}.
\newblock


\bibitem[\protect\citeauthoryear{Zhang, Zheng, and Qi}{Zhang
  et~al\mbox{.}}{2017}]%
        {zhang2017deep}
\bibfield{author}{\bibinfo{person}{Junbo Zhang}, \bibinfo{person}{Yu Zheng},
  {and} \bibinfo{person}{Dekang Qi}.} \bibinfo{year}{2017}\natexlab{}.
\newblock \showarticletitle{Deep spatio-temporal residual networks for citywide
  crowd flows prediction}. In \bibinfo{booktitle}{\emph{Proceedings of
  Thirty-First AAAI Conference on Artificial Intelligence}}
  \emph{(\bibinfo{series}{AAAI '17})}.
\newblock


\bibitem[\protect\citeauthoryear{Zhang, Zheng, Qi, Li, and Yi}{Zhang
  et~al\mbox{.}}{2016}]%
        {DNN_SIG16}
\bibfield{author}{\bibinfo{person}{Junbo Zhang}, \bibinfo{person}{Yu Zheng},
  \bibinfo{person}{Dekang Qi}, \bibinfo{person}{Ruiyuan Li}, {and}
  \bibinfo{person}{Xiuwen Yi}.} \bibinfo{year}{2016}\natexlab{}.
\newblock \showarticletitle{DNN-Based Prediction Model for Spatial-Temporal
  Data}. In \bibinfo{booktitle}{\emph{Proceedings of the 24th ACM SIGSPATIAL
  International Conference on Advances in Geographic Information Systems.}}
  \emph{(\bibinfo{series}{SIGSPATIAL '16})}.
\newblock


\bibitem[\protect\citeauthoryear{Zheng and Ni}{Zheng and Ni}{2013}]%
        {zheng2013time}
\bibfield{author}{\bibinfo{person}{Jiangchuan Zheng} {and}
  \bibinfo{person}{Lionel~M Ni}.} \bibinfo{year}{2013}\natexlab{}.
\newblock \showarticletitle{Time-dependent trajectory regression on road
  networks via multi-task learning}. In
  \bibinfo{booktitle}{\emph{Twenty-Seventh AAAI Conference on Artificial
  Intelligence}}.
\newblock


\end{thebibliography}

\end{document}